\documentclass{article}

\usepackage{iclr2026_conference,times}
\usepackage[utf8]{inputenc}
\usepackage[T1]{fontenc}
\usepackage{hyperref}
\hypersetup{hypertexnames=false}
\usepackage{url}
\usepackage{booktabs}
\usepackage{multirow}
\usepackage{colortbl}   
\usepackage{graphicx}
\usepackage{subcaption}
\usepackage{wrapfig}
\usepackage{amsmath}
\usepackage{amssymb}
\usepackage{amsfonts}
\usepackage{amsthm}
\usepackage{nicefrac}
\usepackage{microtype}
\usepackage{enumitem}
\usepackage{algorithm}
\usepackage{algpseudocode}
\usepackage{tikz}
\usepackage{pgfplots}
\pgfplotsset{compat=1.18}
\usepgfplotslibrary{groupplots}
\usetikzlibrary{positioning,arrows.meta,fit,backgrounds,shapes.geometric,shapes.misc,decorations.pathreplacing,calc,patterns}
\usepackage[section]{placeins}

\definecolor{otblue}{HTML}{0072B2}
\definecolor{otorange}{HTML}{E69F00}
\definecolor{otteal}{HTML}{009E73}
\definecolor{otpurple}{HTML}{CC79A7}
\definecolor{otgray}{HTML}{5A5A5A}

\definecolor{cours}{HTML}{FCCE8B}
\definecolor{cgrpo}{HTML}{9DABC9}
\definecolor{copsd}{HTML}{E8A0A0}
\definecolor{cseed}{HTML}{87C7B3}
\definecolor{crepair}{HTML}{9B6BD1}    
\definecolor{cunguided}{HTML}{6E6E6E}  
\definecolor{cgain}{HTML}{EDAA47}  

\newtheorem{theorem}{Theorem}
\newtheorem{proposition}[theorem]{Proposition}
\newtheorem{lemma}[theorem]{Lemma}

\theoremstyle{definition}
\newtheorem{definition}[theorem]{Definition}

\theoremstyle{remark}

\usepackage{listings}
\usepackage{float}       
\usepackage{textcomp}    

\newcommand{\algsub}[1]{\\[2pt]{\normalfont\footnotesize #1}}

\definecolor{cKeyword}{HTML}{2B3FD9}   
\definecolor{cComment}{HTML}{0E8074}   
\definecolor{cString} {HTML}{E8912D}
\definecolor{cFunc}   {HTML}{D6336C}   
\definecolor{cNet}    {HTML}{8A4FE0}   

\lstdefinestyle{algpy}{
  language=Python,
  morekeywords={yield},
  basicstyle=\ttfamily\footnotesize,
  keywordstyle=\color{cKeyword}\bfseries,
  commentstyle=\color{cComment},
  stringstyle=\color{cString},
  emph={sample,write_note,revise,logp,detach,clip,mean,pick,bad,leaks,
        grpo_loss,sum,len,failed,map},
  emphstyle={\color{cFunc}},
  emph=[2]{pi},
  emphstyle=[2]{\color{cNet}},
  showstringspaces=false,
  columns=fullflexible,
  keepspaces=true,
  upquote=true,
  aboveskip=0.4em, belowskip=0pt,
  xleftmargin=0pt,
}
\newcommand{\methodabbr}{RSHD}
\newcommand{\method}{\textnormal{\textsc{\methodabbr}}}
\newcommand{\methodlong}{Replay-Selected Hindsight Distillation}

\newcommand{\E}{\mathbb{E}}

\newcommand{\D}{\mathcal{D}}          

\newcommand{\loss}{\mathcal{L}}
\newcommand{\KL}{D_{\mathrm{KL}}}
\newcommand{\sg}{\operatorname{sg}}
\newcommand{\clip}{\operatorname{clip}}
\newcommand{\ok}{\operatorname{ok}}

\newcommand{\old}{\mathrm{old}}
\newcommand{\teacher}{\pi_{\mathrm{T}}} 

\newif\ifdrafttodos
\drafttodostrue
\ifdrafttodos
  \usepackage[textsize=small,colorinlistoftodos]{todonotes}
\else
  \usepackage[disable]{todonotes}
\fi

\usepackage{xcolor}
\definecolor{pendingred}{HTML}{C0141B}

\title{Failure Is the Mother of Success: Recovering Supervision with Grounding Hindsight}

\author{Sheldon Yu$^{1}$, Tong Yu$^{2}$, Nikki Lijing Kuang$^{1}$, Zhendong Chu$^{3}$, Zhenwei Tang$^{4}$,\\
\textbf{Sungchul Kim$^{2}$, Lina Yao$^{5}$, Julian McAuley$^{1}$, Junda Wu$^{2}$}\\
{\normalfont $^{1}$University of California, San Diego\quad
$^{2}$Adobe Research\quad
$^{3}$University of Virginia}\\
{\normalfont $^{4}$University of Toronto \quad
$^{5}$University of New South Wales}\\
{\normalfont\small\texttt{\{ziy040,l1kuang,jmcauley\}@ucsd.edu}\quad
\texttt{\{tyu,sukim,jundaw\}@adobe.com}}\\
{\normalfont\small\texttt{zc9uy@virginia.edu}\quad
\texttt{josephtang@cs.toronto.edu}\quad
\texttt{lina.yao@unsw.edu.au}}}

\iclrfinalcopy

\begin{document}
\maketitle
\lhead{Preprint}
\newcommand{\sheldon}[1]{\textcolor{red}{#1}}

\begin{abstract}
Large language models solve reasoning problems by generating long chains of steps, where a single mistake made in the middle steps can negatively impact the final answer. Reinforcement learning with verifiable rewards trains on the
final answer alone: when every attempt at a problem fails there is no successful attempt to
imitate and no spread between attempts to score, so GRPO learns nothing from exactly the problems
it most needs to learn.
The failed reasoning can nevertheless suggest what to
try next. Prior work either reveals the reference answer when it retries, trusts self-written
guidance without testing it, or re-scores the failed attempt instead of sampling a new one. We
propose \methodlong{} (\method{}), which enables \textit{reasoning explorations} from the failure itself. 
Rather than resampling blindly or revealing the answer, the policy reasons over its own note on what went wrong, and
explores from notes that beat the group become supervision during GRPO training. The policy writes a corrective note about each failed
attempt without seeing a reference answer or solution, then re-solves the problem with the
note in context. A note written without the answer cannot be checked for truth, only
for effect: after screening for notes that state the answer, it is admitted only if its exploration replays
succeed more often than the original group did, and an admitted replay is distilled into the
plain-prompt policy. Because admission turns on fresh successes rather than on the failed
rollout, supervision is recovered from groups in which every attempt failed, and from
mixed-group failures alike. Only the policy is
trained, the only judge is the outcome verifier, and deployment requires no notes or extra
inference. Across model families and sizes and two training domains, \method{} consistently
improves over GRPO on in-domain benchmarks, and the gains carry over to out-of-domain
evaluation sets.

\end{abstract}

\section{Introduction}
\label{sec:intro}

\begin{wrapfigure}{r}{0.40\linewidth}
\setlength{\abovecaptionskip}{4pt}\setlength{\belowcaptionskip}{0pt}
\centering
\begin{tikzpicture}
\begin{axis}[
  width=5.3cm, height=3.1cm,
  ybar, bar width=8pt,
  xmin=0.5, xmax=5.5, xtick={1,2,3,4,5},
  ymin=0, ymax=60, ytick={0,20,40,60},
  xlabel={MATH difficulty level}, ylabel={all-fail groups (\%)},
  ymajorgrids=true, grid style={otgray!20, line width=0.3pt},
  axis line style={otgray!60}, axis lines*=left,
  tick style={otgray!60}, major tick length=2pt,
  tick label style={font=\rmfamily\scriptsize, text=black},
  label style={font=\rmfamily\scriptsize, text=black},
  nodes near coords, point meta=explicit symbolic,
  every node near coord/.append style={font=\rmfamily\tiny, text=black},
]
\addplot[fill=cgrpo, draw=cgrpo!70!black] coordinates {
  (1,8.3) [8] (2,24.7) [25] (3,26.9) [27] (4,34.9) [35] (5,51.0) [51]
};
\end{axis}
\end{tikzpicture}
\caption{Groups that GRPO have no learning signal. On the hardest level, half the groups carry no gradient at all.}
\label{fig:allfail-rate}
\end{wrapfigure}
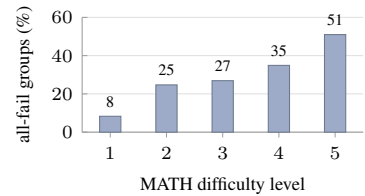

Group-relative policy optimization (GRPO) trains reasoning models by comparing verifier rewards across completions of the same problem \citep{shao2024deepseekmath,deepseekr1,surana2026rollout}. When every completion fails, all group-relative advantages are zero.
These \emph{all-fail groups} provide no task-reward gradient even though their trajectories may reveal recognizable mistakes and unproductive strategies \citep{yu2025stateaware,wei2026reasoningenergy},
which potentially can guide the policy on \textit{reasoning explorations}. 
This is common precisely on hard problems: if a problem is solved with probability $0.1$, an eight-sample group is all-fail more than $40\%$ of the time,
and in our math training run roughly half of the groups at the hardest MATH level carry no gradient in Figure~\ref{fig:allfail-rate}. 
GRPO therefore tends to improve problems the policy already sometimes solves while leaving many failures at the capability frontier unused.


Turning these failures into supervision is difficult. 
Conditioning retries on the reference answer introduces information unavailable at test time \citep{yang2026rlsd}, 
while trusting self-generated feedback without verification can reinforce incorrect diagnoses.
In addition, simply re-scoring a failed rollout can increase the probability of the same mistaken tokens. 
Useful supervision from failure should instead comes from \textit{reasoning explorations}, 
which can reveal a better behavior, verify that it helps, and transfer that behavior back to the plain-prompt policy.

We propose \methodlong{} (\method{}), a failure-conditioned exploration and distillation procedure, which enalbes the \textit{reasoning explorations} in GRPO. 
After the verifier scores a group, the same policy summarizes what went wrong in each failed attempt without access to the reference answer, 
and then re-solves the problem with this note in context. 
A note is retained only when its fresh exploration replays outperform the original group, and successful replays are distilled into the plain-prompt policy. 
Thus, even an all-fail group can yield verified supervision despite having zero GRPO advantage. 
Unlike increasing temperature or nucleus width, which only diversifies retries, 
\method{} conditions exploration on a diagnosis of the specific failed reasoning.
The procedure requires no auxiliary model and adds no inference-time guidance.

Prior work also recovers all-fail signals and internalizes guided retries
\citep{xia2026hill,shi2026erl,chen2026cure,xia2026ovcsd,sleiman2026psd}.
In \method{} the note test is not optional. Across our math run one note in nine states the answer
outright and three in five fail the replay test (Table~\ref{tab:notes}). The test is affordable
because the group GRPO has already sampled supplies the baseline to beat, so the structure that
creates the all-fail problem also pays for its remedy. Section~\ref{sec:related} details the
differences from these methods and from re-scoring the failed attempt \citep{wu2026seed}.

We evaluate \method{} against GRPO across model families and sizes and two training domains,
competition mathematics and free-form science questions, on in-domain and out-of-domain
benchmarks. \method{} improves over GRPO consistently, with the largest gains on the hardest
problems, and the gains carry over out of domain. A sizeable share of the admitted notes come from
all-fail groups, where the original reward signal is zero, and the trained models solve problems
in fewer tokens and run out of budget less often. The extra cost is training-time generation only,
roughly double the tokens sampled per step, and nothing changes at inference.
We summarize our contributions as follows:

\begin{itemize}[leftmargin=*,topsep=2pt,itemsep=1pt]
\item \textbf{Reasoning explorations from failed reasoning.}
We develop a failure-conditioned exploration and distillation procedure that can discover
successful outcomes without a successful exemplar in the initial group. The note writer never sees
the reference answer, and a note is used only after its replays beat the group.
\item \textbf{Replay-selected transfer alongside GRPO.}
\method{} transfers what this exploration finds into the plain-prompt policy alongside GRPO,
keeping replay selection and imitation separate from GRPO's reward normalization.
We characterize the selected-target distribution and the auxiliary signal available when
all original advantages vanish.
\item \textbf{An empirical study of the resulting training recipe.}
We report improved reasoning performance and analyze the notes and replays that supply
supervision, including those recovered from all-fail groups, without adding deployment-time
guidance.
\end{itemize}

\section{Related Work}
\label{sec:related}

\textbf{Reinforcement Learning on LLM Reasoning.}
Reinforcement learning first turned language models into instruction followers by optimizing a
learned preference model with PPO \citep{ouyang2022instructgpt,schulman2017ppo,xie2025pluralistic,li2026importancedpo,surana2026massdpo,kveton2025activedpo,wu2025irpo}. On reasoning the
preference model is unnecessary: an automatic verifier scores the final answer, so the reward is
free of a second trained model \citep{deepseekr1}. GRPO drops the critic as well and estimates an
advantage by comparing the rollouts of one prompt against their own mean
\citep{shao2024deepseekmath}, and subsequent work refines that estimator
\citep{mundada2026wsgrpo,xiong2026rrpo,surana2026fgrpo} and how its loss is
aggregated, decoupling the clipping range and summing over tokens \citep{dapo} or removing the
standard-deviation normalization that biases the gradient \citep{drgrpo}. Exploration in these
methods comes from sampling alone. A higher temperature, a wider nucleus, or updates that keep
the policy's entropy from collapsing \citep{dapo,cui2025entropy,cheng2025entropy} make retries
more varied, but no retry uses why an earlier attempt failed. In all of them the
group is the unit of learning, which is also where they run out: when every rollout of a prompt
earns the same reward the advantages vanish and the prompt contributes nothing. \method{} is
built for that case.

\textbf{Hindsight Exploration.}
Hindsight Experience Replay reinterprets a failed trajectory through the goal it actually reached
\citep{andrychowicz2017her}, and later work settles how much credit a relabeled trajectory should
carry, which relabeled goals to train on, and how to iterate the idea as imitation
\citep{rauber2019hpg,harutyunyan2019hca,zhao2018energyher,ghosh2021gcsl}. In reinforcement
learning the achieved goal is handed back by the environment; in language it must be inferred
from the model's own text, which is where leakage and trust become problems: goals are inferred
from outputs, a failed rationale is regenerated with the answer revealed, or inference is
sidestepped by keeping only verified samples
\citep{zhang2023hir,zelikman2022star,singh2023restem,xia2025sand}. A recent line aims directly at all-fail
reasoning groups, retrying with verbal feedback, training a reflect-and-retry loop, converting
failures into guidance, hinting from the reference, or restoring student states and distilling
verified continuations
\citep{shinn2023reflexion,shi2026erl,wang2026hhd,lv2026failforge,xia2026hill,chen2026cure,xia2026ovcsd,sleiman2026psd,vishe2026skillr1,yu2026olivia}.
\method{} belongs to this family and differs in three ways. Its writer sees neither the reference
answer nor a solution and receives no training objective of its own; a note is admitted by
comparing its replays against the already sampled GRPO group, all-fail groups included; and the
original group keeps its advantages while a uniformly chosen replay is transferred through a
separate plain-prompt loss, with no state restoration or restriction to successful replays.

\begin{figure}[t]
\centering
\includegraphics[width=\linewidth]{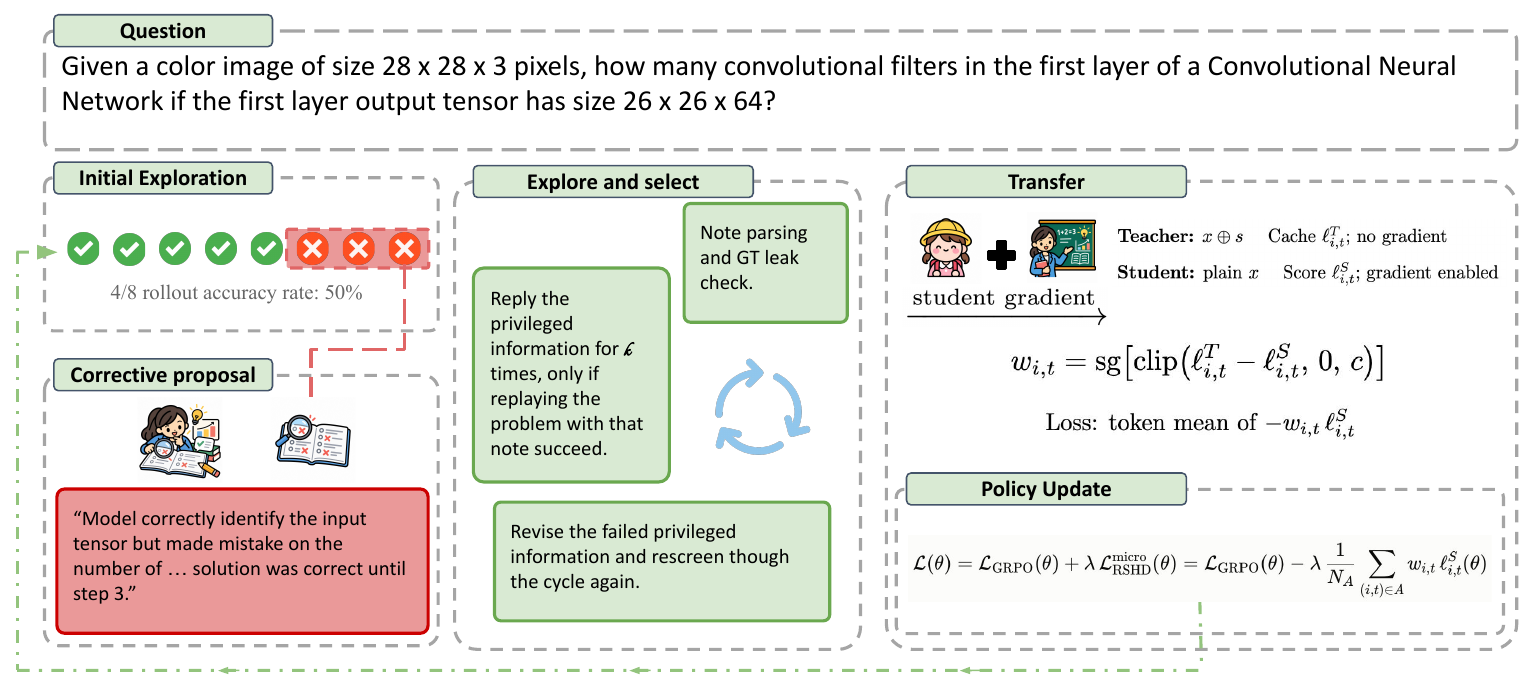}
\caption{\textbf{\method{} overview.} From a scored GRPO group, each failed attempt yields a
corrective note (left). The note is screened for answer leaks and replayed; it is admitted only
if the replays beat the group's own success rate, with one revision round (middle). An admitted
replay is scored by the same weights with and without the note, and the plain-prompt policy
imitates it with detached lift weights, added to the GRPO loss (right).}
\label{fig:main}
\end{figure}

\textbf{On-Policy Self-Distillation with Privileged Context.}
Distillation can transfer privileged information into a policy that lacks it at deployment
\citep{hinton2015distilling,lopezpaz2016privileged,wang2025selfparam}. GKD and on-policy self-distillation
supervise student-generated behavior with a teacher evaluated on the same prefixes
\citep{agarwal2024gkd,zhao2026selfdistilled,hubotter2026sdpo,liu2026sdpg}, and the open design
choice is what the teacher reads. Prior work fills the slot with the reference answer, which
leaks and can destabilize training \citep{yang2026rlsd}, with a successful peer completion from
the same group \citep{li2026hsd}, with a summary of the trajectory \citep{wang2026skillsd,zhou2026filesystem,hu2026skillbrew}, or,
closest to us, with a hindsight skill the policy writes about its own completed trajectory and
re-scores the original actions under, after a skill-SFT stage \citep{wu2026seed}. All of these
keep the sampled trajectory fixed and admit the context on sight. \method{} tests the context
through fresh outcomes before using it, and lets the auxiliary target change: the distilled
tokens are a continuation sampled under the note and selected by its outcome, so the weighted
imitation loss is not standard student-on-policy distillation.

\section{Problem Formulation}
\label{sec:prelim}

\paragraph{Task and notation.}
Let $(x,g)\sim\D$ be a problem and its reference answer, and let $y=(y_1,\ldots,y_T)$ be a
completion, with $y_{<t}$ its prefix before token $t$. A verifier returns $R(x,y)\in\{0,1\}$ by
checking the final answer of $y$ against $g$ and reads nothing else. A prompt policy
$\pi_\theta$ that maximizes
\[
J(\theta)=\E_{(x,g)\sim\D}\,\E_{y\sim\pi_\theta(\cdot\mid x)}\big[R(x,y)\big].
\]

\begin{definition}[Finite-horizon verifier MDP]
\label{def:reasoning-mdp}
Fix a completion budget $H$ for the task and trajectory above, so $T\le H$.
At RL time $t=0,\ldots,H$, the state
$h_t=(x,y_{\le t},H-t)$ records its generated prefix ($y_{\le0}=\varnothing$) and remaining
budget. For $t<H$, an action $a_t$ appends the next token $y_{t+1}$ (or an abstract reasoning
step \citep{wu2025ctrls,wu2025ocean}); an end-of-sequence action enters an absorbing state. The transition is deterministic
and rewards are zero before termination. The only terminal reward is the verifier's
$R(x,y)$ defined above; the reference $g$ is available to that verifier, not to the policy.
Write $G_x=\{y:R(x,y)=1\}$.
\end{definition}
A policy $\pi(a_t\mid h_t)$ induces a trajectory law $P_\pi^x$, with success probability
$J_x(P_\pi^x)=P_\pi^x(G_x)$, so the training objective can equivalently be viewed as maximizing verifier success over induced trajectories.

\paragraph{Group-relative learning and on-policy distillation.}
At each update, a frozen snapshot $\pi_{\theta_{\old}}$ samples a group
$\{y_i\}_{i=1}^n$ for each problem $x$. 
```latex
Let
\begin{equation*}
R_i=R(x,y_i),\qquad \hat p=\frac1n\sum_{i=1}^n R_i,\qquad
F=\{i:R_i=0\},\qquad
\hat A_i=\frac{R_i-\hat p}{\sigma_R+\varepsilon},
\end{equation*}
where $\sigma_R$ is the within-group reward standard deviation and
$\varepsilon>0$ is a numerical stabilizer. GRPO minimizes
\begin{equation}
\loss_{\mathrm{GRPO}}(\theta)
=-\frac1N\sum_{i=1}^{n}\sum_{t=1}^{|y_i|}
\min\!\Big(\rho_{i,t}\hat A_i,\,
\clip(\rho_{i,t},1-\epsilon_{\mathrm{clip}},1+\epsilon_{\mathrm{clip}})\hat A_i\Big),
\label{eq:grpo}
\end{equation}
where
\begin{equation*}
\rho_{i,t}=\frac{\pi_\theta(y_{i,t}\mid x,y_{i,<t})}
{\pi_{\theta_{\old}}(y_{i,t}\mid x,y_{i,<t})},
\qquad N=\sum_i|y_i|.
\end{equation*}
Here $\hat A_i$ is a single rollout-level statistic shared by all tokens of
rollout $i$, unlike the state-action advantage $A_t^\pi(h,a)$. Thus,
\textbf{failed completions reveal neither where the reasoning went wrong nor
how to correct it}. Moreover, if $p_0=J_x(P_{\pi_{\theta_{\old}}}^x)$, an
independently sampled group is all-fail with probability $(1-p_0)^n$, in which
case every $\hat A_i=0$ and the group provides no task-reward gradient.

For on-policy distillation, let $s$ denote privileged context and define the
frozen teacher
\begin{equation*}
\teacher(\cdot)\triangleq\sg[\pi_{\theta_{\old}}](\cdot\mid x\oplus s).
\end{equation*}
For a sequence $u$, teacher and student log-probabilities at the same prefix are
\begin{equation}
\ell_t^{\mathrm T}=\log\teacher(u_t\mid x\oplus s,u_{<t}),\qquad
\ell_t^{\mathrm S}(\theta)=\log\pi_\theta(u_t\mid x,u_{<t}),
\label{eq:teacher-student}
\end{equation}
and the generic distillation objective is
\begin{equation}
\loss_{\mathrm{OPSD}}(\theta;s,u)
=-\sum_t \sg[w_t]\,\ell_t^{\mathrm S}(\theta),
\label{eq:opd}
\end{equation}
where $w_t\ge0$ is a fixed token weight, typically determined from the
teacher--student score gap $\ell_t^{\mathrm T}-\ell_t^{\mathrm S}$.
In \method{}, $u$ is a fresh continuation sampled under privileged context
$s$; teacher scores are stop-gradient and only the student
$\pi_\theta(\cdot\mid x)$ is updated.



\section{Method: \methodlong{}}
\label{sec:method}

A student who gets a problem wrong learns more from working out what went wrong and reasoning over how to explore a different strategy,
rather than being handed the answer, and learns nothing at all from being told only that eight
attempts were wrong. 
\method{} fills the three open slots of Section~\ref{sec:prelim} with that
idea:
\begin{figure}[ht]
\centering
\begin{minipage}[t]{0.48\textwidth}
\begin{algorithm}[H]
\caption{\method{}: explore and select.\algsub{Kept only if its replays beat the group.}}
\label{alg:explore}
\begin{lstlisting}
# R(y): 1 if y solves problem x
# g: answer, used only by leaks()
ys = pi.sample(x, n=8)
p = mean(map(R, ys))
for y in failed(ys):
    s = pi.write_note(x, y)
    for rnd in (1, 2):
        if bad(s) or leaks(s, g):
            break   # screened out
        zs = pi.sample(x+s, n=2)
        r = mean(map(R, zs))
        if r > p:   # admitted
            yield s, pick(zs)
            break
        if rnd == 1:
            s = pi.revise(s,r,zs)
\end{lstlisting}
\end{algorithm}
\end{minipage}\hfill
\begin{minipage}[t]{0.48\textwidth}
\begin{algorithm}[H]
\caption{\method{}: policy update.\algsub{Imitated in proportion to the note's lift.}}
\label{alg:update}
\begin{lstlisting}
# kept: Algorithm 1's (s, replay)
# c: weight cap, lam: aux coef
loss = grpo_loss(ys)  # unchanged
num, ntok = 0.0, 0
for s, yt in kept:
    # one model, scored twice
    lt = pi.logp(yt, x+s).detach()
    ls = pi.logp(yt, x)
    # per-token lift, capped at c
    w = clip(lt-ls, 0, c).detach()
    num += (w * ls).sum()
    ntok += len(yt)  # w=0 too
# nothing kept -> plain GRPO
if ntok > 0:
    loss -= lam * num / ntok
return loss
\end{lstlisting}
\end{algorithm}
\end{minipage}
\end{figure}

a failed attempt is used to write a corrective note, the note is admitted only if solving the problem
again with it in context succeeds more often than the original group did, and a solution
produced under an admitted note is distilled into the plain-prompt policy. 
The original group keeps its GRPO advantages, while the admitted replays add a separate \textit{reasoning exploration} signal. 
Figure~\ref{fig:main} shows the data flow, and Algorithms~\ref{alg:explore} and~\ref{alg:update} the two halves in
pseudocode: one problem is shown, batched in practice.

\subsection{The  Replay-Selected Hindsight Distillation (RSHD) objective}
\label{sec:method-objective}

Start from \eqref{eq:opd} with the privileged context $s_i$ being a note written about the
failed rollout $y_i$, and make two choices. 
First, the sequence that carries the credit is not
$y_i$ but a \emph{replay}: a full solution the same snapshot produced with the note in context, $\tilde y_i\;\sim\;\pi_{\theta_{\old}}(\cdot\mid x\oplus s_i)$.

Re-scoring $y_i$ under the note could only raise the probability of tokens the model already wrote,
the mistaken ones included, whereas a replay contains the tokens the note actually steered. 
Second, each replay token is weighted by the note's detached, clipped, positive \emph{lift} on it, so that credit
concentrates where the note changed the policy and vanishes where it did not. 
Scoring $\tilde y_i$ at identical prefixes under both prompts as in \eqref{eq:teacher-student}, the loss
over the credited positions $\mathcal A$ of a batch is
\begin{equation}
w_{i,t}=\sg\!\Big[\clip\!\big(\ell^{\mathrm T}_{i,t}-\ell^{\mathrm S}_{i,t}(\theta),\,0,\,c\big)\Big], \qquad
\loss_{\method}(\theta)
=-\frac1{|\mathcal A|}\sum_{(i,t)\in\mathcal A}w_{i,t}\,\ell^{\mathrm S}_{i,t}(\theta),
\label{eq:gh}
\end{equation}
where $c$ caps the weight so that no token dominates, and
\begin{equation*}
\mathcal A\triangleq\big\{(i,t):i\in\mathcal U,\ 1\le t\le|\tilde y_i|\big\},
\qquad
\loss(\theta)=\loss_{\mathrm{GRPO}}(\theta)+\lambda\,\loss_{\method}(\theta),
\end{equation*}
where $\mathcal A$ contains every token of every admitted replay, including zero-weight positions, and $\mathcal U$ is the admitted set defined in Section~\ref{sec:method-exploration}.
with only failed rollouts contributing to the second term; successful ones are left to GRPO.
Three properties follow directly. \textbf{Non-empty updates under failure.} On an all-fail group every advantage in \eqref{eq:grpo} is zero, while any admitted replay with positive weight gives \eqref{eq:gh} a nonzero gradient. \textbf{Exact fallback to GRPO.} When no note is admitted, $\mathcal A$ is empty, the second term vanishes, and the step reduces to plain GRPO. \textbf{No deployment overhead.} 
Because the note appears only in the teacher prompt, deployment uses the plain prompt with no extra inference, while the outcome-selected targets and teacher-preference weights make \eqref{eq:gh} distinct from standard student-on-policy distillation and place trust in the note-admission procedure.

\subsection{Reasoning exploration and note admission}
\label{sec:method-exploration}

\textbf{Reasoning exploration.} For each $i\in F$ the snapshot writes a short corrective note $s_i$
from $(x,y_i,\text{``incorrect''})$: what went wrong and what to do instead, at most 80 words,
returned as a structured lesson.
The writer sees neither the reference answer nor a solution, so notes need no
successful exemplar.
Groups the policy solves every time are skipped, and failures in mixed and
all-fail groups are treated alike. 
A note is worth using when a fresh attempt is likelier to
succeed with it than without it. Writing
\begin{equation*}
p_0\;\triangleq\;\E_{y\sim\pi_{\theta_{\old}}(\cdot\mid x)}\big[R(x,y)\big],
\qquad
p_s\;\triangleq\;\E_{\tilde y\sim\pi_{\theta_{\old}}(\cdot\mid x\oplus s)}\big[R(x,\tilde y)\big],
\end{equation*}
the note helps exactly when $p_s>p_0$. Both are properties of fresh attempts rather than of the
note's text, and neither is observable, so we estimate the comparison by replay.

\textbf{Screening and admitting.} Two checks run before a single replay is sampled. Write
$\ok(s)=1$ when $s$ parses and does not state the reference answer (a string match of the
normalized $g$ against the note text), and $\ok(s)=0$ otherwise; a note failing either check is
discarded at no sampling cost. A note with $\ok(s_i)=1$ is then used to solve the problem $k$
times under the rollout sampling settings, and admitted iff its replays beat the original group,
\begin{equation}
\hat r(s_i)=\frac1k\sum_{j=1}^{k}R\big(x,\tilde y^{(j)}\big),\qquad
\mathrm{keep}(s_i)\iff\hat r(s_i)>\hat p ,
\label{eq:keep}
\end{equation}
so that the admitted set of a problem is
\begin{equation*}
    \mathcal U\;\triangleq\;\big\{\,i\in F\ :\ \ok(s_i)=1\ \text{ and }\ \hat r(s_i)>\hat p\,\big\}.
\end{equation*}
A note that falls short is revised once: the writer sees the note, its tally, and excerpts of up
to three failed replays, again without the reference, and the revision is screened and tested
against the same $\hat p$; a second failure ends the search. For $i\in\mathcal U$ the target in
\eqref{eq:gh} is one of that note's replays drawn uniformly,
$\tilde y_i\sim\mathrm{Unif}\{\tilde y^{(1)},\ldots,\tilde y^{(k)}\}$, whether or not it succeeded.


\textbf{What admission guarantees.} Admission in \eqref{eq:keep} provides enrichment rather than purity: for an all-fail group, a note is admitted if at least one of its $k$ replays succeeds, and a uniformly sampled admitted replay is correct with probability
\begin{equation*}
\Pr\!\big[R(x,\tilde y_i)=1\mid \hat r(s_i)>0\big]
=\frac{p_s}{1-(1-p_s)^k}\ge p_s,
\end{equation*}
so the distilled replay is more likely to be correct but is not guaranteed to be correct. Because an answer-revealing note could satisfy this test trivially, the leak screen is applied before admission, while experiments separately report note survival, replay improvement, and distilled-replay correctness.

\section{Failure-Conditioned Exploration and Transfer}
\label{sec:exploration-theory}

In the verifier MDP of Definition~\ref{def:reasoning-mdp}, two failures motivate \textit{reasoning exploration}. \textbf{No reward contrast:} an all-fail GRPO group, which occurs with probability $(1-p_0)^n$ (Section~\ref{sec:prelim}), provides no task-reward signal (Lemma~\ref{lem:allfail-signal}). \textbf{Local repair ceiling:} a failed trajectory may enter a strategy from which success is unreachable within the remaining budget \citep{yu2026selfloops}, while \textbf{restarting with failure-derived context} can explore a different strategy. We formalize the second failure and the restart below; proofs appear in Appendix~\ref{app:exploration-theory}.


\noindent\textbf{No reward contrast in an all-fail group.}
Let $P_0=P_{\pi_{\theta_{\old}}}^x$ be the rollout distribution of original prompt and
$p_0=J_x(P_0)$ its population success rate. There are no advantages if rollouts are all fail:

\begin{lemma}[All-fail GRPO reward signal]
\label{lem:allfail-signal}
For $n$ independent $Y_i\sim P_0$, define the event
$\mathcal E_{\mathrm{fail}}=\{R(x,Y_i)=0\text{ for every }i\}$. Then
\begin{equation*}
    \Pr(\mathcal E_{\mathrm{fail}}\mid x)=(1-p_0)^n.
\end{equation*}
On $\mathcal E_{\mathrm{fail}}$, every group-relative advantage $\hat A_i$ is
zero and the task-reward part of this group's clipped GRPO surrogate
has zero gradient.
\end{lemma}

This is a finite-sample loss of reward signal, 
while KL regularization and other problems can still supply gradients.
In-rollout self-correction can append reasoning, but cannot erase an earlier prefix or
recover spent budget.

\label{sec:synthetic-results}
\newcommand{\synthHardAucHigh}{+5.1\,\mathrm{pp}}
\newcommand{\synthHardAucHighCI}{1.3\,\mathrm{pp}}
\newcommand{\synthHardAucLow}{-7.1\,\mathrm{pp}}
\newcommand{\synthHardAucLowCI}{0.8\,\mathrm{pp}}
\newcommand{\synthHardSwitchAuc}{42.5\pm1.3}
\newcommand{\synthHardUnguidedAuc}{31.6\pm3.4}

\newcommand{\synthRepairAUC}{0.7}
\newcommand{\synthRestartAUC}{42.5}
\newcommand{\synthNeutralAdmit}{0.9}
\newcommand{\synthSwitchAdmit}{4.1}
\newcommand{\synthNeutralSelected}{50.2}
\newcommand{\synthSwitchSelected}{50.7}
\newcommand{\synthCoupledHarm}{56.1}
\newcommand{\synthCoupledHarmCI}{4.9}
\newcommand{\synthCoupledCertified}{3.3}
\newcommand{\synthEscapeRepairBiased}{+2.3\pm1.3}
\newcommand{\synthEscapeBiasedFrequent}{+22.4\pm1.3}
\newcommand{\synthEscapeFrequentFull}{+17.1\pm2.3}

\begin{figure}[t]
\centering
\begin{tikzpicture}
\begin{groupplot}[
  group style={group size=3 by 1, horizontal sep=0.85cm, ylabels at=edge left},
  width=0.36\linewidth, height=0.27\linewidth,
  ymode=log, ymin=0.004, ymax=1.15, ytick={0.01,0.1,1}, yticklabels={0.01,0.1,1},
  xmin=-5, xmax=125, xtick={0,40,80,120}, xlabel={training update},
  grid=major, grid style={otgray!15}, axis line style={otgray!60},
  axis lines*=left, major tick length=1.5pt,
  tick label style={font=\rmfamily\scriptsize, text=black},
  label style={font=\rmfamily\footnotesize, text=black, align=center},
  title style={font=\rmfamily\footnotesize, yshift=-2pt},
  legend style={font=\rmfamily\footnotesize, draw=none, fill=none, legend columns=4, /tikz/every even column/.append style={column sep=0.45cm}},
  every axis plot/.append style={line width=0.9pt, mark size=1.3pt, mark repeat=2},
]
\nextgroupplot[title={budget $H=3$}, ylabel={plain-prompt success}, legend to name=synthLegend]
\addplot[draw=none, fill=cgrpo, fill opacity=0.22, forget plot] coordinates {(0,0.008912)(10,0.01147)(20,0.0154)(30,0.02153)(40,0.03378)(50,0.07084)(60,0.3072)(70,0.8424)(80,0.9611)(90,0.9655)(100,0.9731)(110,0.9791)(120,0.9825)(120,0.9778)(110,0.9705)(100,0.9527)(90,0.9094)(80,0.6699)(70,0.3587)(60,0.1185)(50,0.03853)(40,0.02166)(30,0.01542)(20,0.01211)(10,0.009932)(0,0.008399)} -- cycle;
\addplot[draw=none, fill=crepair, fill opacity=0.22, forget plot] coordinates {(0,0.008912)(10,0.01817)(20,0.04598)(30,0.2185)(40,0.7152)(50,0.9162)(60,0.9547)(70,0.9698)(80,0.9773)(90,0.9823)(100,0.9851)(110,0.9877)(120,0.9892)(120,0.9871)(110,0.9853)(100,0.9824)(90,0.9788)(80,0.9709)(70,0.9581)(60,0.9299)(50,0.7938)(40,0.3257)(30,0.07033)(20,0.02568)(10,0.01275)(0,0.008399)} -- cycle;
\addplot[draw=none, fill=cunguided, fill opacity=0.22, forget plot] coordinates {(0,0.008912)(10,0.03078)(20,0.1927)(30,0.8159)(40,0.9339)(50,0.9623)(60,0.9737)(70,0.9804)(80,0.984)(90,0.987)(100,0.9886)(110,0.9903)(120,0.9914)(120,0.9903)(110,0.9893)(100,0.9877)(90,0.9858)(80,0.9826)(70,0.9783)(60,0.9703)(50,0.9551)(40,0.9103)(30,0.6227)(20,0.1118)(10,0.02192)(0,0.008399)} -- cycle;
\addplot[draw=none, fill=cours, fill opacity=0.22, forget plot] coordinates {(0,0.008912)(10,0.04987)(20,0.4641)(30,0.8651)(40,0.9404)(50,0.9638)(60,0.9738)(70,0.98)(80,0.9835)(90,0.9866)(100,0.9883)(110,0.9899)(120,0.991)(120,0.9899)(110,0.9889)(100,0.9874)(90,0.9854)(80,0.9824)(70,0.9786)(60,0.9715)(50,0.9602)(40,0.9322)(30,0.831)(20,0.3266)(10,0.03663)(0,0.008399)} -- cycle;
\addplot[color=cgrpo, mark=triangle*, solid] coordinates {(0,0.008655)(10,0.0107)(20,0.01376)(30,0.01848)(40,0.02772)(50,0.05469)(60,0.2128)(70,0.6006)(80,0.8155)(90,0.9374)(100,0.9629)(110,0.9748)(120,0.9802)};
\addlegendentry{GRPO}
\addplot[color=crepair, mark=pentagon*, densely dotted] coordinates {(0,0.008655)(10,0.01546)(20,0.03583)(30,0.1444)(40,0.5204)(50,0.855)(60,0.9423)(70,0.964)(80,0.9741)(90,0.9806)(100,0.9838)(110,0.9865)(120,0.9882)};
\addlegendentry{repair-biased restart ($\epsilon=0.1$)}
\addplot[color=cunguided, mark=square*, densely dashed] coordinates {(0,0.008655)(10,0.02635)(20,0.1523)(30,0.7193)(40,0.9221)(50,0.9587)(60,0.972)(70,0.9793)(80,0.9833)(90,0.9864)(100,0.9882)(110,0.9898)(120,0.9908)};
\addlegendentry{unguided retry}
\addplot[color=cours, mark=*, solid] coordinates {(0,0.008655)(10,0.04325)(20,0.3953)(30,0.848)(40,0.9363)(50,0.962)(60,0.9726)(70,0.9793)(80,0.983)(90,0.986)(100,0.9878)(110,0.9894)(120,0.9905)};
\addlegendentry{strategy-switch note}
\nextgroupplot[title={budget $H=5$}]
\addplot[draw=none, fill=cgrpo, fill opacity=0.22, forget plot] coordinates {(0,0.006678)(10,0.007594)(20,0.008712)(30,0.00988)(40,0.01134)(50,0.01301)(60,0.01586)(70,0.01936)(80,0.025)(90,0.03373)(100,0.05564)(110,0.1144)(120,0.3143)(120,0.1394)(110,0.06321)(100,0.03767)(90,0.02639)(80,0.0205)(70,0.0164)(60,0.01359)(50,0.01158)(40,0.009944)(30,0.008797)(20,0.007877)(10,0.007196)(0,0.006478)} -- cycle;
\addplot[draw=none, fill=crepair, fill opacity=0.22, forget plot] coordinates {(0,0.006678)(10,0.01037)(20,0.01528)(30,0.02462)(40,0.04516)(50,0.1102)(60,0.303)(70,0.5758)(80,0.8265)(90,0.917)(100,0.9422)(110,0.9582)(120,0.9668)(120,0.9555)(110,0.9381)(100,0.8994)(90,0.7981)(80,0.5581)(70,0.2373)(60,0.06332)(50,0.03196)(40,0.02227)(30,0.01473)(20,0.01057)(10,0.008009)(0,0.006478)} -- cycle;
\addplot[draw=none, fill=cunguided, fill opacity=0.22, forget plot] coordinates {(0,0.006678)(10,0.01413)(20,0.03421)(30,0.09674)(40,0.3579)(50,0.7564)(60,0.8976)(70,0.938)(80,0.9568)(90,0.967)(100,0.9735)(110,0.9783)(120,0.9812)(120,0.9787)(110,0.9748)(100,0.9692)(90,0.9614)(80,0.9467)(70,0.9173)(60,0.8346)(50,0.5604)(40,0.1991)(30,0.06144)(20,0.02527)(10,0.01139)(0,0.006478)} -- cycle;
\addplot[draw=none, fill=cours, fill opacity=0.22, forget plot] coordinates {(0,0.006678)(10,0.01845)(20,0.05905)(30,0.2377)(40,0.6472)(50,0.8533)(60,0.9202)(70,0.9467)(80,0.9607)(90,0.969)(100,0.9745)(110,0.9787)(120,0.9814)(120,0.9795)(110,0.9763)(100,0.9717)(90,0.9659)(80,0.9564)(70,0.939)(60,0.9044)(50,0.8158)(40,0.5586)(30,0.1725)(20,0.04426)(10,0.01539)(0,0.006478)} -- cycle;
\addplot[color=cgrpo, mark=triangle*, solid] coordinates {(0,0.006578)(10,0.007395)(20,0.008295)(30,0.009339)(40,0.01064)(50,0.0123)(60,0.01472)(70,0.01788)(80,0.02275)(90,0.03006)(100,0.04665)(110,0.08882)(120,0.2269)};
\addplot[color=crepair, mark=pentagon*, densely dotted] coordinates {(0,0.006578)(10,0.009188)(20,0.01293)(30,0.01968)(40,0.03372)(50,0.07107)(60,0.1832)(70,0.4065)(80,0.6923)(90,0.8575)(100,0.9208)(110,0.9482)(120,0.9611)};
\addplot[color=cunguided, mark=square*, densely dashed] coordinates {(0,0.006578)(10,0.01276)(20,0.02974)(30,0.07909)(40,0.2785)(50,0.6584)(60,0.8661)(70,0.9277)(80,0.9518)(90,0.9642)(100,0.9714)(110,0.9766)(120,0.9799)};
\addplot[color=cours, mark=*, solid] coordinates {(0,0.006578)(10,0.01692)(20,0.05165)(30,0.2051)(40,0.6029)(50,0.8346)(60,0.9123)(70,0.9429)(80,0.9586)(90,0.9674)(100,0.9731)(110,0.9775)(120,0.9804)};
\nextgroupplot[title={budget $H=8$}]
\addplot[draw=none, fill=cgrpo, fill opacity=0.22, forget plot] coordinates {(0,0.004507)(10,0.00481)(20,0.005142)(30,0.00544)(40,0.005771)(50,0.006057)(60,0.006498)(70,0.006961)(80,0.007288)(90,0.007775)(100,0.008462)(110,0.009346)(120,0.01031)(120,0.008707)(110,0.008049)(100,0.007447)(90,0.007026)(80,0.006627)(70,0.006212)(60,0.005923)(50,0.005584)(40,0.005381)(30,0.005101)(20,0.004806)(10,0.004559)(0,0.004345)} -- cycle;
\addplot[draw=none, fill=crepair, fill opacity=0.22, forget plot] coordinates {(0,0.004507)(10,0.005394)(20,0.006612)(30,0.007993)(40,0.01054)(50,0.01314)(60,0.01754)(70,0.02296)(80,0.03102)(90,0.04515)(100,0.07561)(110,0.1445)(120,0.2819)(120,0.06456)(110,0.03961)(100,0.03016)(90,0.02235)(80,0.01717)(70,0.0137)(60,0.01144)(50,0.009106)(40,0.007758)(30,0.006441)(20,0.005527)(10,0.00481)(0,0.004345)} -- cycle;
\addplot[draw=none, fill=cunguided, fill opacity=0.22, forget plot] coordinates {(0,0.004507)(10,0.00707)(20,0.01228)(30,0.02128)(40,0.03565)(50,0.06569)(60,0.1417)(70,0.3069)(80,0.5754)(90,0.781)(100,0.8721)(110,0.9143)(120,0.9359)(120,0.9129)(110,0.8722)(100,0.7789)(90,0.5953)(80,0.3433)(70,0.1605)(60,0.07559)(50,0.03883)(40,0.02282)(30,0.0143)(20,0.009277)(10,0.005849)(0,0.004345)} -- cycle;
\addplot[draw=none, fill=cours, fill opacity=0.22, forget plot] coordinates {(0,0.004507)(10,0.007881)(20,0.015)(30,0.0288)(40,0.06458)(50,0.1591)(60,0.3748)(70,0.6342)(80,0.7928)(90,0.8689)(100,0.9073)(110,0.9308)(120,0.9453)(120,0.9384)(110,0.9221)(100,0.8944)(90,0.8469)(80,0.7495)(70,0.5538)(60,0.2903)(50,0.1179)(40,0.05247)(30,0.02477)(20,0.01326)(10,0.007165)(0,0.004345)} -- cycle;
\addplot[color=cgrpo, mark=triangle*, solid] coordinates {(0,0.004426)(10,0.004685)(20,0.004974)(30,0.00527)(40,0.005576)(50,0.00582)(60,0.00621)(70,0.006586)(80,0.006957)(90,0.007401)(100,0.007954)(110,0.008697)(120,0.009509)};
\addplot[color=crepair, mark=pentagon*, densely dotted] coordinates {(0,0.004426)(10,0.005102)(20,0.00607)(30,0.007217)(40,0.009149)(50,0.01112)(60,0.01449)(70,0.01833)(80,0.0241)(90,0.03375)(100,0.05288)(110,0.09206)(120,0.1732)};
\addplot[color=cunguided, mark=square*, densely dashed] coordinates {(0,0.004426)(10,0.00646)(20,0.01078)(30,0.01779)(40,0.02924)(50,0.05226)(60,0.1086)(70,0.2337)(80,0.4593)(90,0.6882)(100,0.8255)(110,0.8933)(120,0.9244)};
\addplot[color=cours, mark=*, solid] coordinates {(0,0.004426)(10,0.007523)(20,0.01413)(30,0.02679)(40,0.05852)(50,0.1385)(60,0.3326)(70,0.594)(80,0.7712)(90,0.8579)(100,0.9009)(110,0.9264)(120,0.9419)};
\end{groupplot}
\node[anchor=south] at ($(group c2r1.north)+(0,0.55cm)$) {\pgfplotslegendfromname{synthLegend}};
\end{tikzpicture}
\caption{\textbf{Learning with failure-derived notes in a controlled two-strategy task.} Plain-prompt
success during training (log scale, mean and 95\% interval over ten seeds) at three reasoning budgets $H$,
with initial viable-strategy probability $0.01$ and note strength $1.6$. Every arm is allotted two extra
attempts per rollout. Unguided retry imitates accepted retries without a note,
repair-biased restart proposes a strategy switch with probability $0.1$ and a local repair otherwise, and
the strategy-switch note always proposes the other strategy. Success is measured without a note.}
\label{fig:synthetic-exploration}
\end{figure}
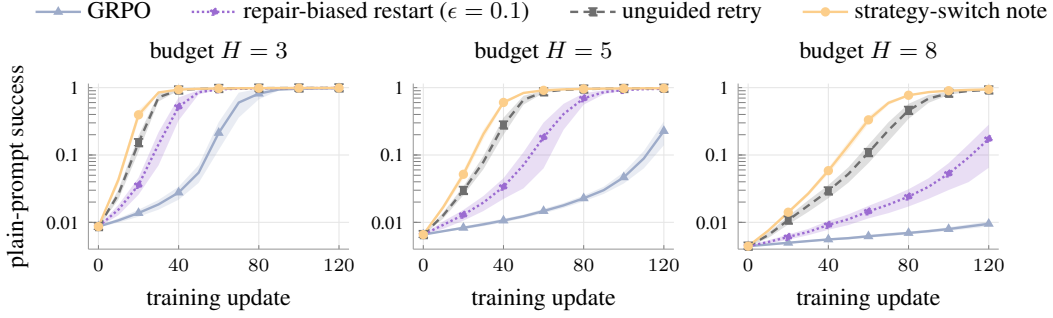


\noindent\textbf{The ceiling of local repair.}
We condition on a fixed problem $x$ and write $P_0=P_{\pi_{\theta_{\old}}}^x$ for its rollout
law, so that $p_0=J_x(P_0)$.

\begin{definition}[Repair basin and escape rate]
\label{def:repair-basin}
For a failed attempt $y^-$, let $\mathcal B(y^-)$ comprise complete, budget-feasible
trajectories preserving its committed prefix or strategy, with
$\mathcal B(y^-)\cap G_x=\varnothing$. A possibly history-dependent reflection kernel
$K_j$ has escape bound $\epsilon$ if, given \emph{every} preceding history
$\mathcal H_{j-1}$ without success,
\begin{equation*}
\Pr_{Y_j\sim K_j(\cdot\mid\mathcal H_{j-1})}
 (Y_j\notin\mathcal B(y^-)\mid\mathcal H_{j-1})\le\epsilon
\quad\text{almost surely}.
\end{equation*}
\end{definition}

The basin captures a locally reasonable but unreachable approach, not a local optimum of the
parameter objective.

\begin{proposition}[Repair-restricted exploration ceiling]
\label{prop:repair-ceiling}
If a repair kernel has escape bound $\epsilon$ (Definition~\ref{def:repair-basin}), then the
chance of success within $m$ retries is at most
\begin{equation*}
    1-(1-\epsilon)^m\le m\epsilon.
\end{equation*}
Prefix- or strategy-preserving repair has $\epsilon=0$ and cannot succeed in this case.
\end{proposition}

The restriction is on the retry \emph{kernel}, not the name of a method: reflection that restarts
from the problem and changes strategy need not have small $\epsilon$
\citep{shinn2023reflexion,shi2026erl}. The proposition identifies when error repair alone
cannot reach a solution; Appendix~\ref{app:exploration-theory} gives a horizon-3 witness.

\noindent\textbf{Restarting under a failure-derived proposal.}
After observing a failure, the same policy writes a context $s$ and solves again from the
original problem.

\begin{definition}[Failure-conditioned contextual proposal]
\label{def:contextual-proposal}
The current snapshot writes a context $s$ from a failed rollout and its incorrect
verdict, then produces fresh complete attempts under
$T_s=P_{\pi_{\theta_{\old}}}^{x\oplus s}$. Conditional on the original
history and fixed $s$, define
\begin{equation*}
p_s=T_s(G_x),\qquad
e_s=T_s\bigl(\mathcal B(y^-)^c\bigr).
\end{equation*}
The escape rate $e_s$ measures a change of strategy; only $p_s$ measures its
outcome. Replays are on-policy for this snapshot on the augmented prompt, not for
the deployed plain-prompt policy.
\end{definition}

Choosing $s$ selects a trajectory-scale proposal without changing model weights, analogous
to a temporally extended policy choice \citep{sutton1999options}. It may expose an existing
but rarely sampled strategy; the note need not identify an erroneous step.

\begin{proposition}[Guided discovery and selected replays]
\label{prop:replay-selection}
Draw $k\ge1$ independent complete replays from $T_s$. Admit $s$ if at least one succeeds,
and choose one replay uniformly from the accepting batch, with law $Q_s$.
For $p_s>0$,
\[
\Pr(\text{admit }s\mid s)=1-(1-p_s)^k,\qquad
J_x(Q_s)=\frac{p_s}{1-(1-p_s)^k}\ge p_s.
\]
If $p_s>p_0$, guided replays discover success more often than $k$ plain retries; if
$p_s>\epsilon$, they exceed the same-budget repair ceiling of
Proposition~\ref{prop:repair-ceiling}.
\end{proposition}

On an all-fail group the method's replay-rate test reduces to this any-success event.
Admission does not prove $p_s>p_0$: a neutral note can pass by chance. Population uplift
is an empirical premise, while the selected target can still be incorrect.

\noindent\textbf{Transferring discovery into the plain policy.}
The original GRPO advantages are unchanged. An OPSD-style auxiliary imitates a selected
replay without its note. Let $\pi_{\mathrm G}$ and $\pi_{\mathrm A}$ be the prompt
policies after the GRPO-only \emph{local} update and after adding that auxiliary,
with trajectory laws $P_{\mathrm G}=P_{\pi_{\mathrm G}}^x$ and
$P_{\mathrm A}=P_{\pi_{\mathrm A}}^x$. 
The next result states what transfer must achieve, rather than assuming that any token-weighted imitation
objective achieves it.

\begin{theorem}[Conditional local non-degradation]
\label{thm:local-transfer}
Suppose the admitted replay target has law $Q$ with population margin
$J_x(Q)\ge J_x(P_{\mathrm G})+\delta$ for $\delta>0$. For $\alpha\in[0,1]$ let
$M_\alpha=(1-\alpha)P_{\mathrm G}+\alpha Q$ be a mixture of \emph{complete trajectories}.
If $\operatorname{TV}(P_{\mathrm A},M_\alpha)\le\xi$, then
\begin{equation}
J_x(P_{\mathrm A})-J_x(P_{\mathrm G})\ge\alpha\delta-\xi.
\label{eq:local-transfer-bound}
\end{equation}
Thus $\xi\le\alpha\delta$ suffices for non-degradation on $x$. If no replay is admitted and
no auxiliary update is taken, the local update equals GRPO's.
\end{theorem}

This is a conservative policy-improvement condition \citep{kakade2002approximately,schulman2015trpo},
not a guarantee supplied by the current positive-lift loss: selected replay sampling and
per-token weighting do not certify closeness to $M_\alpha$. 
The proof and a token-level KL sufficient condition appear in
Appendix~\ref{app:exploration-theory}. In the simplest case, however, the loss does move the
successful root decision:

\begin{proposition}[Two-strategy detached-lift witness]
\label{prop:strategy-lift}
Consider a standalone one-decision problem with strategies $0,1$ and
$R=1$ exactly for $1$. The original group contains only $0$.
Let the plain policy have $p=\sigma(\phi)\in(0,1)$ and the fixed-context
teacher have $q\in(p,1)$, where $\sigma(z)=(1+e^{-z})^{-1}$.
Assume there are no other gradients on $\phi$. From $k$ independent
teacher replays admit any batch containing $1$ and choose one replay
uniformly. For $c,\eta,\lambda>0$, take one gradient step on
$-\lambda w_z\log\pi_\phi(z)$ with detached weights
\begin{equation*}
w_1=\min\{\log(q/p),c\}>0,\qquad
w_0=\max\{0,\min\{\log((1-q)/(1-p)),c\}\}=0.
\end{equation*}
Conditional on admission, the plain success rate never decreases on
this local step and increases strictly in expectation:
\begin{equation*}
\E[p_{\mathrm{new}}\mid E_s]
=\mu\,\sigma\bigl(\phi+\eta\lambda w_1(1-p)\bigr)
 +(1-\mu)p>p,\qquad
\mu=\frac{q}{1-(1-q)^k}.
\end{equation*}
The original group's task-reward GRPO gradient is zero.
\end{proposition}

The witness depends on credit being placed on the successful root decision
and on that logit being unaffected by other trajectories. With shared
parameters, multiple decisions, or a different auxiliary objective,
positive guided success or a nonzero imitation gradient alone cannot
guarantee improvement of the plain policy.

\textbf{A controlled illustration.}
Figure~\ref{fig:synthetic-exploration} runs these steps in a small verifier MDP with two strategies.
One is locally valid but cannot finish within the budget $H$, and the other succeeds only if several
later decisions are also correct. A single two-logit policy produces both the GRPO group and the
replays, and a note shifts one of its logits by a fixed amount. GRPO barely improves at $H=8$, where
almost every group fails. At $H=5$ and $H=8$ the strategy-switch note learns fastest, while a restart
that leaves the failed strategy only one time in ten lags far behind at $H=8$, as the repair ceiling
of Proposition~\ref{prop:repair-ceiling} predicts. Unguided retry with the same extra budget comes
close by the end of training ($0.92$ against $0.94$ at $H=8$), so here the note mainly speeds up
discovery. The note effects are fixed by the model rather than written by a policy, so the figure
shows what happens when a note helps, not that real notes do. 
\section{Experiments}
\label{sec:experiments}

\textbf{Setup.}
\label{sec:setup}
Backbones are Qwen3-1.7B, Qwen3-4B and Qwen3-8B \citep{qwen3}, three sizes spanning roughly a
factor of five in parameters, used as released. Each is trained twice. The math domain trains on the 7,498 MATH
training problems \citep{hendrycks2021math} and is evaluated on MATH-500
\citep{lightman2024verify}. The science domain trains on 2,141 MMLU-Pro questions
\citep{wang2024mmlupro} with numeric answers, asked free-form, and is evaluated on 300 questions
held out from that pool, the free-form GPQA-Diamond set of \citet{chandak2025answermatching}
(126 questions), and the 619 numeric questions of TheoremQA \citep{chen2023theoremqa}; the latter
two are disjoint from the pool. In Table~\ref{tab:main} the MATH-500 column reports the
math-trained models and the three science columns the science-trained models.

\textbf{Metric.}
Accuracy averaged over 8 samples per problem (temperature $0.6$, 4,096 tokens), graded by an
exact verifier. GPQA-Diamond, whose answers are often phrases, carries a second number in which
rejected answers are re-judged by the evaluated model with a yes/no prompt
(Appendix~\ref{app:template}); the matcher never enters training.

\textbf{Baselines.}
\emph{Base} is the released model. All trained rows share the GRPO backbone, rollouts, and step
budget and differ only in the added loss. \emph{OPSD} \citep{zhao2026selfdistilled} re-scores
the student's rollouts under the reference solution; \emph{SEED} \citep{wu2026seed} re-scores
them under a self-written note kept without testing (our writer and prompts, without its
skill-SFT stage). \method{} differs from SEED in testing the note and from both in distilling a
continuation sampled under it rather than the rollout.

\textbf{Training.}
All baseline use 150 steps of 14 problems $\times$ 8 rollouts, learning rate $10^{-6}$, DAPO
token-level loss, and 3,072-token completions. \method{} uses $k=2$ replays, two note rounds,
$\lambda=0.2$, $c=2$, and no KL penalty. Full settings are in Appendix~\ref{app:impl}.

\begin{table}[t]
\centering
\small
\begin{tabular}{llcccc}
\toprule
Backbone & Method & MATH-500 & MMLU-Pro  & GPQA-Diamond & TheoremQA \\
\midrule
\multirow{5}{*}{Qwen3-4B}
 & Base & 65.2 & 30.2 & 1.8\,/\,3.1 & 30.6 \\
 & GRPO & 76.9 & 48.7 & 3.0\,/\,9.6 & 40.1 \\
 & OPSD & 73.6 & 34.5 & 3.1\,/\,6.2 & 34.8 \\
 & SEED & 75.2 & 41.7 & 2.3\,/\,7.5 & 38.7 \\
\rowcolor{cyan!10} & \method{} (ours) & \textbf{84.5} & \textbf{61.2} & \textbf{9.1\,/\,27.0} & \textbf{56.5} \\
\midrule
\multirow{5}{*}{Qwen3-1.7B}
 & Base & 64.2 & 27.3 & 0.3\,/\,2.8 & 25.3 \\
 & GRPO & 68.1 & 31.1 & 1.6\,/\,5.3 & 28.4 \\
 & OPSD & 70.3 & 30.9 & 1.3\,/\,3.7 & 28.8 \\
 & SEED & \textbf{71.2} & 29.8 & 1.3\,/\,8.7 & 28.4 \\
\rowcolor{cyan!10} & \method{} (ours) & 70.0 & \textbf{40.4} & \textbf{4.9\,/\,17.2} & \textbf{40.8} \\
\midrule
\multirow{5}{*}{Qwen3-8B}
 & Base & 62.3 & 28.0 & 1.2\,/\,2.4 & 26.8 \\
 & GRPO & 75.3 & 45.1 & 2.8\,/\,8.7 & 35.2 \\
 & OPSD & 71.0 & 32.2 & 2.7\,/\,4.1 & 31.9 \\
 & SEED & 74.4 & 42.8 & 3.1\,/\,12.7 & 38.6 \\
\rowcolor{cyan!10} & \method{} (ours) & \textbf{80.6} & \textbf{60.7} & \textbf{8.9\,/\,22.7} & \textbf{55.9} \\
\bottomrule
\end{tabular}
\caption{\textbf{\method{} improves over GRPO on every benchmark and every backbone.} Accuracy
(\%, mean over 8 samples per problem); GPQA-Diamond shows exact\,/\,matched. The MATH-500
column reports models trained on MATH, and the other three report models trained on the MMLU-Pro
pool, so GPQA-Diamond and TheoremQA are out of domain. All trained rows use the same GRPO rollouts
and 150-step budget. Best in bold.}
\label{tab:main}
\end{table}

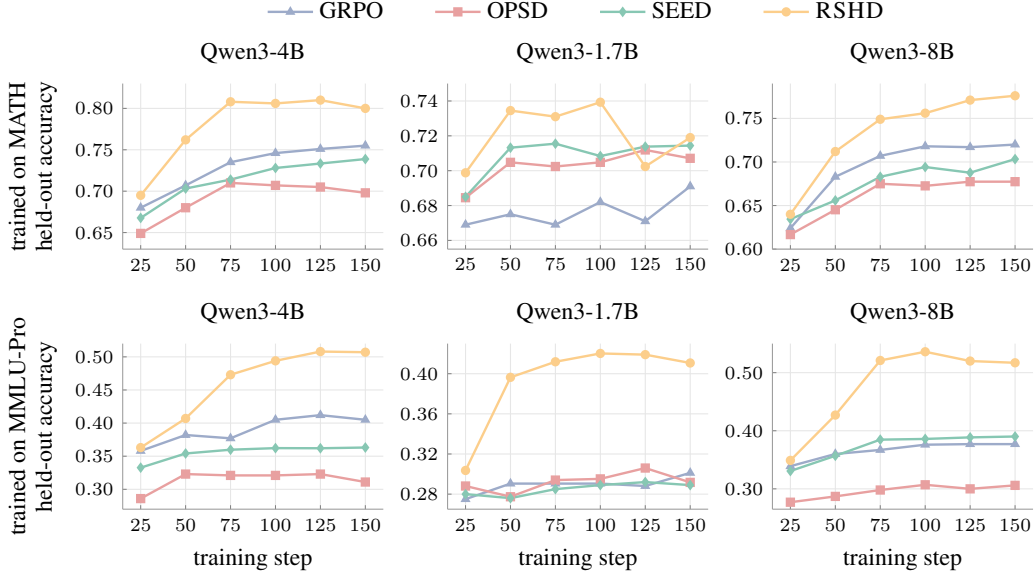
\begin{figure}[ht]
\centering
\begin{tikzpicture}
\begin{groupplot}[
  group style={group size=3 by 2, horizontal sep=0.85cm, vertical sep=1.25cm,
               xlabels at=edge bottom, ylabels at=edge left},
  width=0.36\linewidth, height=0.27\linewidth,
  xmin=15, xmax=160, xtick={25,50,75,100,125,150},
  xlabel={training step},  
  yticklabel style={/pgf/number format/fixed, /pgf/number format/fixed zerofill, /pgf/number format/precision=2},
  grid=major, grid style={otgray!15}, axis line style={otgray!60},
  axis lines*=left,  
  major tick length=1.5pt,  
  tick label style={font=\rmfamily\scriptsize, text=black},
  label style={font=\rmfamily\footnotesize, text=black, align=center},
  title style={font=\rmfamily\footnotesize, yshift=-2pt},
  legend style={font=\rmfamily\footnotesize, draw=none, fill=none, legend columns=4, /tikz/every even column/.append style={column sep=0.6cm}},
  every axis plot/.append style={line width=0.9pt, mark size=1.3pt},
]
\nextgroupplot[title={Qwen3-4B}, ymin=0.63, ymax=0.83, ytick={0.65,0.70,0.75,0.80},
              ylabel={trained on MATH\\held-out accuracy},
              legend to name=curvesLegend]  
\addplot[color=cgrpo, mark=triangle*] coordinates {(25,0.680)(50,0.707)(75,0.735)(100,0.746)(125,0.751)(150,0.755)};
\addlegendentry{GRPO}
\addplot[color=copsd, mark=square*] coordinates {(25,0.649)(50,0.680)(75,0.710)(100,0.707)(125,0.705)(150,0.698)};
\addlegendentry{OPSD}
\addplot[color=cseed, mark=diamond*] coordinates {(25,0.6677)(50,0.7032)(75,0.7140)(100,0.7280)(125,0.7334)(150,0.7388)};
\addlegendentry{SEED}
\addplot[color=cours, mark=*] coordinates {(25,0.695)(50,0.762)(75,0.808)(100,0.806)(125,0.810)(150,0.800)};
\addlegendentry{\method{}}
\nextgroupplot[title={Qwen3-1.7B}, ymin=0.655, ymax=0.75, ytick={0.66,0.68,0.70,0.72,0.74}]
\addplot[color=cgrpo, mark=triangle*] coordinates {(25,0.669)(50,0.675)(75,0.669)(100,0.682)(125,0.671)(150,0.691)};
\addplot[color=copsd, mark=square*] coordinates {(25,0.6845)(50,0.7048)(75,0.7024)(100,0.7048)(125,0.7119)(150,0.7071)};
\addplot[color=cseed, mark=diamond*] coordinates {(25,0.6850)(50,0.7132)(75,0.7155)(100,0.7084)(125,0.7138)(150,0.7144)};
\addplot[color=cours, mark=*] coordinates {(25,0.6988)(50,0.7345)(75,0.7310)(100,0.7393)(125,0.7024)(150,0.7190)};
\nextgroupplot[title={Qwen3-8B}, ymin=0.60, ymax=0.79, ytick={0.60,0.65,0.70,0.75}]
\addplot[color=cgrpo, mark=triangle*] coordinates {(25,0.624)(50,0.683)(75,0.707)(100,0.718)(125,0.717)(150,0.720)};
\addplot[color=copsd, mark=square*] coordinates {(25,0.6167)(50,0.6452)(75,0.6750)(100,0.6726)(125,0.6774)(150,0.6774)};
\addplot[color=cseed, mark=diamond*] coordinates {(25,0.6344)(50,0.6559)(75,0.6830)(100,0.6942)(125,0.6877)(150,0.7033)};
\addplot[color=cours, mark=*] coordinates {(25,0.640)(50,0.712)(75,0.749)(100,0.756)(125,0.771)(150,0.776)};
\nextgroupplot[title={Qwen3-4B}, ymin=0.27, ymax=0.52, ytick={0.30,0.35,0.40,0.45,0.50},
              ylabel={trained on MMLU-Pro\\held-out accuracy}]
\addplot[color=cgrpo, mark=triangle*] coordinates {(25,0.358)(50,0.382)(75,0.377)(100,0.405)(125,0.412)(150,0.405)};
\addplot[color=copsd, mark=square*] coordinates {(25,0.286)(50,0.323)(75,0.321)(100,0.321)(125,0.323)(150,0.311)};
\addplot[color=cseed, mark=diamond*] coordinates {(25,0.3327)(50,0.3541)(75,0.3598)(100,0.3622)(125,0.3620)(150,0.3631)};
\addplot[color=cours, mark=*] coordinates {(25,0.363)(50,0.407)(75,0.473)(100,0.494)(125,0.508)(150,0.507)};
\nextgroupplot[title={Qwen3-1.7B}, ymin=0.265, ymax=0.43, ytick={0.28,0.32,0.36,0.40}]
\addplot[color=cgrpo, mark=triangle*] coordinates {(25,0.2750)(50,0.2905)(75,0.2905)(100,0.2905)(125,0.2881)(150,0.3012)};
\addplot[color=copsd, mark=square*] coordinates {(25,0.2881)(50,0.2774)(75,0.2940)(100,0.2952)(125,0.3060)(150,0.2917)};
\addplot[color=cseed, mark=diamond*] coordinates {(25,0.280)(50,0.276)(75,0.285)(100,0.289)(125,0.292)(150,0.289)};
\addplot[color=cours, mark=*] coordinates {(25,0.3036)(50,0.3964)(75,0.4119)(100,0.4202)(125,0.4190)(150,0.4107)};
\nextgroupplot[title={Qwen3-8B}, ymin=0.265, ymax=0.55, ytick={0.30,0.40,0.50}]
\addplot[color=cgrpo, mark=triangle*] coordinates {(25,0.339)(50,0.360)(75,0.367)(100,0.376)(125,0.377)(150,0.377)};
\addplot[color=copsd, mark=square*] coordinates {(25,0.277)(50,0.287)(75,0.298)(100,0.307)(125,0.300)(150,0.306)};
\addplot[color=cseed, mark=diamond*] coordinates {(25,0.3307)(50,0.3569)(75,0.3847)(100,0.3859)(125,0.3886)(150,0.3901)};
\addplot[color=cours, mark=*] coordinates {(25,0.349)(50,0.427)(75,0.521)(100,0.536)(125,0.520)(150,0.517)};
\end{groupplot}
\node[anchor=south] at ($(group c2r1.north)+(0,0.55cm)$) {\pgfplotslegendfromname{curvesLegend}};
\end{tikzpicture}
\caption{\textbf{Held-out accuracy during training.} Rows are the training domain, named on the
$y$-axis (top: MATH, held out on 100 MATH-500 problems; bottom: the MMLU-Pro numeric pool, held
out on 100 of its 300 reserved questions), columns are the backbone; every 25 steps, 8 samples
per problem at the training temperature. Each panel's $y$-range is fitted to its own curves, so compare gaps within a panel rather
than across panels. \method{} separates from GRPO by step 50 on all three backbones, and the gap
is widest on the science domain, where GRPO plateaus early.}
\label{fig:curves}
\end{figure}

\subsection{Main results}
\label{sec:exp_main}

Table~\ref{tab:main} is the main comparison. On MATH, \method{} reaches 84.5 on MATH-500 against
76.9 for GRPO at 4B, a gain of 7.6 points from the same 150 steps and the same rollouts, and 80.6
against 75.3 at 8B. Section~\ref{sec:exp_abl} breaks the 4B result down by problem difficulty.
The gain comes with a change in behavior: \method{} writes shorter solutions (1,979
tokens on average against 2,545), runs out of budget half as often (13.2\% against 26.2\%), and
leaves a problem without a boxed answer half as often (8.9\% against 18.2\%). The share of
problems solved on all 8 samples rises from 66.4\% to 72.6\% and pass@8 from 86.2\% to 92.0\%,
so the improvement is consistent as well as in the mean.

On the science domain the gains are larger. On the held-out slice of the training pool
\method{} reaches 61.2 against 48.7 at 4B and 60.7 against 45.1 at 8B, and the two external sets
move with it: TheoremQA rises from 40.1 to 56.5 at 4B and from 35.2 to 55.9 at 8B, and
GPQA-Diamond, where every model is weak, from 9.6 to 27.0 matched at 4B. The science-trained
models also improve on MATH-500, which they never trained on (74.2 to 81.3 at 4B, 70.7 to 77.5 at
8B), so what the notes teach is not confined to the domain they were written about.

\textbf{Training dynamics.}
Figure~\ref{fig:curves} shows the held-out accuracy during training for every backbone and
training domain. On the 4B and 8B backbones the two curves nearly coincide at step 25 and separate by
step 50, after which \method{} stays ahead at every checkpoint: the final gap on MATH is 4.5
points at 4B and 5.6 at 8B, and on the science pool about 10 points at 4B and 14 at 8B, where
GRPO plateaus within the first 75 steps while \method{} keeps improving until step 100.

\subsection{Ablations}
\label{sec:exp_abl}

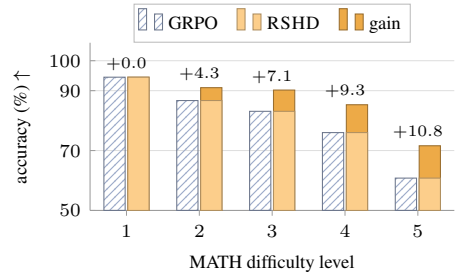
\begin{wrapfigure}{r}{0.47\linewidth}
\setlength{\abovecaptionskip}{4pt}\setlength{\belowcaptionskip}{0pt}
\centering
\begin{tikzpicture}
\begin{axis}[
  width=6.4cm, height=3.8cm,
  ybar, bar width=8pt,
  xmin=0.5, xmax=5.5, xtick={1,2,3,4,5},
  ymin=50, ymax=106, ytick={50,70,90,100},
  xlabel={MATH difficulty level}, ylabel={accuracy (\%) $\uparrow$},
  ymajorgrids=true, grid style={otgray!20, line width=0.3pt},
  axis line style={otgray!60}, axis lines*=left,
  tick style={otgray!60}, major tick length=2pt,
  tick label style={font=\rmfamily\scriptsize, text=black},
  label style={font=\rmfamily\scriptsize, text=black},
  legend style={at={(0.5,1.0)}, anchor=south, legend columns=3,
                draw=otgray!40, fill=white, inner sep=1.5pt, row sep=0pt,
                font=\rmfamily\scriptsize, text=black,
                /tikz/every even column/.append style={column sep=5pt}},
]
\addplot[bar shift=-4.5pt, pattern=north east lines, pattern color=cgrpo, draw=cgrpo!70!black]
  coordinates {(1,94.5)(2,86.7)(3,83.1)(4,76.0)(5,60.8)};
\addlegendentry{GRPO}
\addplot[bar shift=4.5pt, fill=cgain, draw=cgain!65!black, forget plot]
  coordinates {(1,94.5)(2,91.0)(3,90.2)(4,85.3)(5,71.6)};
\addplot[bar shift=4.5pt, fill=cours, draw=cours!70!black]
  coordinates {(1,94.5)(2,86.7)(3,83.1)(4,76.0)(5,60.8)};
\addlegendentry{\method{}}
\addlegendimage{fill=cgain, draw=cgain!65!black}
\addlegendentry{gain}
\node[font=\rmfamily\tiny, text=black] at (axis cs:1,99.0) {$+0.0$};
\node[font=\rmfamily\tiny, text=black] at (axis cs:2,95.5) {$+4.3$};
\node[font=\rmfamily\tiny, text=black] at (axis cs:3,94.7) {$+7.1$};
\node[font=\rmfamily\tiny, text=black] at (axis cs:4,89.8) {$+9.3$};
\node[font=\rmfamily\tiny, text=black] at (axis cs:5,76.1) {$+10.8$};
\end{axis}
\end{tikzpicture}
\caption{Accuracy by MATH's own difficulty label (Qwen3-4B); the $y$-axis starts at $40$. The
darker top segment of each \method{} bar is its gain over GRPO, also printed above the pair.}
\label{fig:level}
\end{wrapfigure}

\textbf{The gain grows with problem difficulty.}
Every MATH problem carries a difficulty label from Level 1 to Level 5, transcribed by
\citet{hendrycks2021math} from the competition sources the problems were drawn from, and MATH-500
inherits it \citep{lightman2024verify}. We did not define the levels, no part of training or note
admission sees them, and their ordering was fixed long before this work, so the split is not one
we chose. Figure~\ref{fig:level} reads the Qwen3-4B math result along it. The two models are indistinguishable at Level 1 ($94.5$ both),
where 43 problems sit near ceiling and a few samples separate the arms. From there the margin
rises monotonically with difficulty, reaching $+10.8$ points on the 134 Level-5 problems. \method{} adds
nothing where GRPO already succeeds and the most where it does not, which is the profile a method
that manufactures signal out of failures should have.

\textbf{Problems the policy cannot solve at all.}
Difficulty is a property of the problem; the condition \method{} is built for is a property of the
group the policy produced. Section~\ref{sec:method-objective} claims the two terms divide the work
exactly at that line: when all $n$ rollouts of a group fail, every advantage in \eqref{eq:grpo} is
zero and GRPO's update is empty, while an admitted replay still carries gradient.
\begin{wraptable}{r}{0.47\linewidth}
\setlength{\tabcolsep}{4pt}
\caption{MATH-500 split by what the \emph{released} Gemma-4-E4B solves in 8 samples. The gain sits
entirely on the problems it never solves. $\Delta$ is computed before rounding.}
\label{tab:allfail}
\footnotesize
\centering
\begin{tabular}{l c || c c | c}
\toprule
Base model & \multirow{2}{*}{$n$} & \multicolumn{2}{c|}{accuracy (\%) $\uparrow$} & \multirow{2}{*}{$\Delta$} \\
solves it & & GRPO & \method{} & \\
\midrule
never ($0/8$)        & 66  & 6.4           & \textbf{15.0} & $+8.5$ \\
sometimes ($1$--$7$) & 172 & \textbf{84.8} & 84.4          & $-0.4$ \\
\midrule
avg                  &  & 45.6          & \textbf{49.7} & $+4.1$ \\
\bottomrule
\end{tabular}
\end{wraptable}

The level
breakdown above is consistent with that claim without testing it, so we test it directly on
Gemma-4-E4B: we stratify MATH-500 by what the \emph{released} model solves in 8 samples, a third
arm that neither trained model can influence, and read the two trained models within each stratum.
The 66 problems the base model never solves are those that would have produced an all-fail group
had they been in training. 
There \method{} reaches $15.0$ against GRPO's $6.4$
(Table~\ref{tab:allfail}); a paired bootstrap over those problems puts the gap at $+8.5$ points
with a $95\%$ interval of $[+4.5,+12.9]$, and \method{} is ahead on 21 of them against 3 the other
way. On the remaining 434 problems the two arms are level to within half a point. The overall
$+0.7$ is therefore not a thin gain spread across the benchmark; it is a large gain on the eighth
of it where GRPO has no signal at all, and a wash everywhere else.

\section{Conclusion}
\label{sec:conclusion}

GRPO receives no learning signal when every rollout in a group fails, leaving many difficult problems underused during training. We introduced RSHD, which turns failed rollouts into corrective notes, validates them through fresh replays, and distills only guidance that improves outcomes. Across model scales and both mathematics and science domains, RSHD consistently outperforms GRPO, with the largest gains on harder problems and no additional inference-time overhead. These results show that failed rollouts are not merely wasted samples: when their lessons are verified before training, they can become a useful source of supervision for improving reasoning models.
\subsection*{Use of large language models}
In this work, we used generative AI tools to implement analysis code, to
provide feedback on the experimental design and on the interpretation of
results, and to help draft the paper, its figures and its LaTeX. 

\bibliographystyle{iclr2026_conference}
\bibliography{references}

@misc{dapo,
  title={DAPO: An Open-Source LLM Reinforcement Learning System at Scale},
  author={Yu, Qiying and Zhang, Zheng and Zhu, Ruofei and Yuan, Yufeng and Zuo, Xiaochen and Yue, Yu and Fan, Tiantian and Liu, Gaohong and Liu, Lingjun and Liu, Xin and others},
  year={2025},
  eprint={2503.14476},
  archivePrefix={arXiv},
  primaryClass={cs.LG}
}

@misc{cui2025entropy,
  title={The Entropy Mechanism of Reinforcement Learning for Reasoning Language Models},
  author={Cui, Ganqu and Zhang, Yuchen and Chen, Jiacheng and Yuan, Lifan and Wang, Zhi and Zuo, Yuxin and Li, Haozhan and Fan, Yuchen and Chen, Huayu and Chen, Weize and Liu, Zhiyuan and Peng, Hao and Bai, Lei and Ouyang, Wanli and Cheng, Yu and Zhou, Bowen and Ding, Ning},
  year={2025},
  eprint={2505.22617},
  archivePrefix={arXiv},
  primaryClass={cs.LG}
}

@misc{cheng2025entropy,
  title={Reasoning with Exploration: An Entropy Perspective},
  author={Cheng, Daixuan and Huang, Shaohan and Zhu, Xuekai and Dai, Bo and Zhao, Wayne Xin and Zhang, Zhenliang and Wei, Furu},
  year={2025},
  eprint={2506.14758},
  archivePrefix={arXiv},
  primaryClass={cs.CL}
}

@misc{drgrpo,
  title={Understanding R1-Zero-Like Training: A Critical Perspective},
  author={Liu, Zichen and Chen, Changyu and Li, Wenjun and Qi, Penghui and Pang, Tianyu and Du, Chao and Lee, Wee Sun and Lin, Min},
  year={2025},
  eprint={2503.20783},
  archivePrefix={arXiv},
  primaryClass={cs.LG}
}

@inproceedings{agarwal2024gkd,
  title={On-Policy Distillation of Language Models: Learning from Self-Generated Mistakes},
  author={Agarwal, Rishabh and Vieillard, Nino and Zhou, Yongchao and Stanczyk, Piotr and Ramos, Sabela and Geist, Matthieu and Bachem, Olivier},
  booktitle={International Conference on Learning Representations},
  year={2024}
}

@article{andrychowicz2017her,
  title={Hindsight Experience Replay},
  author={Andrychowicz, Marcin and Wolski, Filip and Ray, Alex and Schneider, Jonas and Fong, Rachel and Welinder, Peter and McGrew, Bob and Tobin, Josh and Abbeel, Pieter and Zaremba, Wojciech},
  journal={Advances in Neural Information Processing Systems},
  year={2017}
}

@misc{wang2026skillsd,
  title={Skill-SD: Skill-Conditioned Self-Distillation for Multi-turn LLM Agents},
  author={Wang, Hao and Wang, Guozhi and Xiao, Han and Zhou, Yufeng and Pan, Yue and Wang, Jichao and Xu, Ke and Wen, Yafei and Ruan, Xiaohu and Chen, Xiaoxin and Qi, Honggang},
  year={2026},
  eprint={2604.10674},
  archivePrefix={arXiv}
}

@misc{zhao2026selfdistilled,
  title={Self-Distilled Reasoner: On-Policy Self-Distillation for Large Language Models},
  author={Zhao, Siyan and Xie, Zhihui and Liu, Mengchen and Huang, Jing and Pang, Guan and Chen, Feiyu and Grover, Aditya},
  year={2026},
  eprint={2601.18734},
  archivePrefix={arXiv}
}

@misc{liu2026sdpg,
  title={Self-Distilled Policy Gradient},
  author={Liu, Yifeng and Zhang, Shiyuan and Zhang, Yifan and Gu, Quanquan},
  year={2026},
  eprint={2606.04036},
  archivePrefix={arXiv}
}

@article{hinton2015distilling,
  title={Distilling the Knowledge in a Neural Network},
  author={Hinton, Geoffrey and Vinyals, Oriol and Dean, Jeff},
  journal={arXiv preprint arXiv:1503.02531},
  year={2015}
}

@inproceedings{lopezpaz2016privileged,
  title={Unifying Distillation and Privileged Information},
  author={Lopez-Paz, David and Bottou, Leon and Scholkopf, Bernhard and Vapnik, Vladimir},
  booktitle={International Conference on Learning Representations},
  year={2016}
}

@misc{shao2024deepseekmath,
  title={DeepSeekMath: Pushing the Limits of Mathematical Reasoning in Open Language Models},
  author={Shao, Zhihong and Wang, Peiyi and Zhu, Qihao and Xu, Runxin and Song, Junxiao and Bi, Xiao and Zhang, Haowei and Zhang, Mingchuan and Li, Y. K. and Wu, Y. and Guo, Daya},
  year={2024},
  eprint={2402.03300},
  archivePrefix={arXiv}
}

@inproceedings{hendrycks2021math,
  title={Measuring Mathematical Problem Solving with the MATH Dataset},
  author={Hendrycks, Dan and Burns, Collin and Kadavath, Saurav and Arora, Akul and Basart, Steven and Tang, Eric and Song, Dawn and Steinhardt, Jacob},
  booktitle={NeurIPS Datasets and Benchmarks},
  year={2021}
}

@inproceedings{zhang2023hir,
  title={The Wisdom of Hindsight Makes Language Models Better Instruction Followers},
  author={Zhang, Tianjun and Liu, Fangchen and Wong, Justin and Abbeel, Pieter and Gonzalez, Joseph E.},
  booktitle={International Conference on Machine Learning},
  year={2023},
  eprint={2302.05206},
  archivePrefix={arXiv}
}

@misc{singh2023restem,
  title={Beyond Human Data: Scaling Self-Training for Problem-Solving with Language Models},
  author={Singh, Avi and Co-Reyes, John D. and Agarwal, Rishabh and Anand, Ankesh and Patil, Piyush and Garcia, Xavier and Liu, Peter J. and Harrison, James and Lee, Jaehoon and Xu, Kelvin and others},
  year={2023},
  eprint={2312.06585},
  archivePrefix={arXiv}
}

@misc{yang2026rlsd,
  title={Self-Distilled RLVR},
  author={Yang, Chenxu and Qin, Chuanyu and Si, Qingyi and Chen, Minghui and Gu, Naibin and Yao, Dingyu and Lin, Zheng and Wang, Weiping and Wang, Jiaqi and Duan, Nan},
  year={2026},
  eprint={2604.03128},
  archivePrefix={arXiv}
}

@misc{schulman2017ppo,
  title={Proximal Policy Optimization Algorithms},
  author={Schulman, John and Wolski, Filip and Dhariwal, Prafulla and Radford, Alec and Klimov, Oleg},
  year={2017},
  eprint={1707.06347},
  archivePrefix={arXiv}
}

@inproceedings{ouyang2022instructgpt,
  title={Training Language Models to Follow Instructions with Human Feedback},
  author={Ouyang, Long and Wu, Jeffrey and Jiang, Xu and Almeida, Diogo and Wainwright, Carroll and Mishkin, Pamela and Zhang, Chong and Agarwal, Sandhini and Slama, Katarina and Ray, Alex and Schulman, John and Hilton, Jacob and Kelton, Fraser and Miller, Luke and Simens, Maddie and Askell, Amanda and Welinder, Peter and Christiano, Paul and Leike, Jan and Lowe, Ryan},
  booktitle={Advances in Neural Information Processing Systems},
  year={2022}
}

@misc{li2026hsd,
  title={Localizing Credit at the Divergence: Path-Conditioned Self-Distillation for LLM Reasoning},
  author={Li, Yu and Hong, Shu and Lan, Tian},
  year={2026},
  eprint={2606.15576},
  archivePrefix={arXiv},
  primaryClass={cs.LG}
}

@misc{wang2026hhd,
  title={Hindsight Hint Distillation: Scaffolded Reasoning for SWE Agents from CoT-free Answers},
  author={Wang, Shengjie and Li, Guanghe and Yang, Zonghan and Gao, Yang},
  year={2026},
  eprint={2605.11556},
  archivePrefix={arXiv},
  primaryClass={cs.AI}
}

@article{deepseekr1,
  title={DeepSeek-R1 incentivizes reasoning in LLMs through reinforcement learning},
  author={{DeepSeek-AI} and Guo, Daya and Yang, Dejian and Zhang, Haowei and Song, Junxiao and others},
  journal={Nature},
  volume={645},
  pages={633--638},
  year={2025},
  doi={10.1038/s41586-025-09422-z}
}

@inproceedings{lightman2024verify,
  title={Let's Verify Step by Step},
  author={Lightman, Hunter and Kosaraju, Vineet and Burda, Yura and Edwards, Harri and Baker, Bowen and Lee, Teddy and Leike, Jan and Schulman, John and Sutskever, Ilya and Cobbe, Karl},
  booktitle={International Conference on Learning Representations},
  year={2024},
  eprint={2305.20050},
  archivePrefix={arXiv}
}

@misc{hubotter2026sdpo,
  title={Reinforcement Learning via Self-Distillation},
  author={H{\"u}botter, Jonas and others},
  year={2026},
  eprint={2601.20802},
  archivePrefix={arXiv},
  primaryClass={cs.LG}
}

@misc{xia2026ovcsd,
  title={From Scoring to Acting: Outcome-Verified Comparative Self-Distillation for {LLM} Agents},
  author={Xia, Xu and Piao, Jinghua and Yang, Min and Lan, Xiaochong and Chen, Jiaju and Li, Yong},
  year={2026},
  eprint={2607.27937},
  archivePrefix={arXiv},
  primaryClass={cs.LG}
}

@inproceedings{zelikman2022star,
  title={{STaR}: Bootstrapping Reasoning with Reasoning},
  author={Zelikman, Eric and Wu, Yuhuai and Mu, Jesse and Goodman, Noah D.},
  booktitle={Advances in Neural Information Processing Systems},
  year={2022},
  eprint={2203.14465},
  archivePrefix={arXiv}
}

@inproceedings{rauber2019hpg,
  title={Hindsight policy gradients},
  author={Rauber, Paulo and Ummadisingu, Avinash and Mutz, Filipe and Schmidhuber, J{\"u}rgen},
  booktitle={International Conference on Learning Representations (ICLR)}, year={2019}, note={arXiv:1711.06006}}

@inproceedings{zhao2018energyher,
  title={Energy-based hindsight experience prioritization},
  author={Zhao, Rui and Tresp, Volker},
  booktitle={Conference on Robot Learning (CoRL)}, year={2018}, note={arXiv:1810.01363}}

@inproceedings{ghosh2021gcsl,
  title={Learning to reach goals via iterated supervised learning},
  author={Ghosh, Dibya and Gupta, Abhishek and Reddy, Ashwin and Fu, Justin and Devin, Coline and Eysenbach, Benjamin and Levine, Sergey},
  booktitle={International Conference on Learning Representations (ICLR)}, year={2021}, note={arXiv:1912.06088}}

@inproceedings{harutyunyan2019hca,
  title={Hindsight credit assignment},
  author={Harutyunyan, Anna and Dabney, Will and Mesnard, Thomas and Azar, Mohammad and Piot, Bilal and Heess, Nicolas and van Hasselt, Hado and Wayne, Greg and Singh, Satinder and Precup, Doina and Munos, R{\'e}mi},
  booktitle={Advances in Neural Information Processing Systems (NeurIPS)}, year={2019}, note={arXiv:1912.02503}}

@misc{wu2026seed,
  title={{SEED}: Self-Evolving On-Policy Distillation for Agentic Reinforcement Learning},
  author={Wu, Jinyang and Yang, Shuo and Lu, Zhengxi and Zhang, Fan and Shen, Yuhao and Feng, Lang and Luo, Haoran and Lian, Zheng and Zhang, Shuai and Wen, Zhengqi and Tao, Jianhua},
  year={2026},
  eprint={2607.14777},
  archivePrefix={arXiv}
}

@inproceedings{shinn2023reflexion,
  title={Reflexion: Language Agents with Verbal Reinforcement Learning},
  author={Shinn, Noah and Cassano, Federico and Gopinath, Ashwin and Narasimhan, Karthik and Yao, Shunyu},
  booktitle={Advances in Neural Information Processing Systems},
  year={2023}
}

@misc{qwen3,
  title={Qwen3 Technical Report},
  author={Yang, An and others},
  year={2025},
  eprint={2505.09388},
  archivePrefix={arXiv}
}

@inproceedings{wang2024mmlupro,
  title={{MMLU-Pro}: A More Robust and Challenging Multi-Task Language Understanding Benchmark},
  author={Wang, Yubo and Ma, Xueguang and Zhang, Ge and Ni, Yuansheng and Chandra, Abhranil and Guo, Shiguang and Ren, Weiming and Arulraj, Aaran and He, Xuan and Jiang, Ziyan and Li, Tianle and Ku, Max and Wang, Kai and Zhuang, Alex and Fan, Rongqi and Yue, Xiang and Chen, Wenhu},
  booktitle={Advances in Neural Information Processing Systems, Datasets and Benchmarks Track},
  year={2024}
}

@inproceedings{chen2023theoremqa,
  title={{TheoremQA}: A Theorem-driven Question Answering Dataset},
  author={Chen, Wenhu and Yin, Ming and Ku, Max and Lu, Pan and Wan, Yixin and Ma, Xueguang and Xu, Jianyu and Wang, Xinyi and Xia, Tony},
  booktitle={Proceedings of the 2023 Conference on Empirical Methods in Natural Language Processing},
  year={2023}
}

@misc{chandak2025answermatching,
  title={Answer Matching Outperforms Multiple Choice for Language Model Evaluation},
  author={Chandak, Nikhil and Goel, Shashwat and Prabhu, Ameya and Hardt, Moritz and Geiping, Jonas},
  year={2025},
  eprint={2507.02856},
  archivePrefix={arXiv}
}

@misc{xia2026hill,
  title={Learning to Hint for Reinforcement Learning},
  author={Xia, Yu and Xu, Canwen and Yao, Zhewei and McAuley, Julian and He, Yuxiong},
  year={2026},
  eprint={2604.00698},
  archivePrefix={arXiv},
  url={https://arxiv.org/abs/2604.00698}
}

@misc{shi2026erl,
  title={Experiential Reinforcement Learning},
  author={Shi, Taiwei and Chen, Sihao and Jiang, Bowen and Song, Linxin and Yang, Longqi and Zhao, Jieyu},
  year={2026},
  eprint={2602.13949},
  archivePrefix={arXiv},
  url={https://arxiv.org/abs/2602.13949}
}

@inproceedings{chen2026cure,
  title={{CURE}: Critique-Driven Unified Reinforcement Learning for Test-Time Self-Improvement},
  author={Chen, Guirong and Ye, Shuqi and Yang, Wenkai and Shen, Shiqi and Shen, Guangyao and Lin, Yankai},
  booktitle={Proceedings of the 64th Annual Meeting of the Association for Computational Linguistics (Volume 1: Long Papers)},
  year={2026},
  pages={28632--28653},
  publisher={Association for Computational Linguistics},
  doi={10.18653/v1/2026.acl-long.1321},
  url={https://aclanthology.org/2026.acl-long.1321/}
}

@misc{sleiman2026psd,
  title={Privileged On-Policy Self-Distillation},
  author={Sleiman, Essam},
  year={2026},
  month=jul,
  howpublished={Canvas research write-up},
  url={https://www.canvas.inc/research/privileged-self-distillation}
}

@misc{lv2026failforge,
  title={{FailForge}: Distilling Procedural Competence from Persistent Failures into Code Agents},
  author={Lv, Dongyi and E, Fushun and Cai, Aichen and Huang, Liang and Zhang, Ya and Ding, Qiuyu and Wu, Canhui and Wang, Zhi and Zhang, Yuesong and Wang, Jiaqi and Duan, Nan},
  year={2026},
  eprint={2608.08570},
  archivePrefix={arXiv},
  url={https://arxiv.org/abs/2608.08570}
}

@inproceedings{kakade2002approximately,
  title={Approximately Optimal Approximate Reinforcement Learning},
  author={Kakade, Sham and Langford, John},
  booktitle={Proceedings of the Nineteenth International Conference on Machine Learning},
  year={2002}
}

@inproceedings{schulman2015trpo,
  title={Trust Region Policy Optimization},
  author={Schulman, John and Levine, Sergey and Moritz, Philipp and Jordan, Michael and Abbeel, Pieter},
  booktitle={Proceedings of the 32nd International Conference on Machine Learning},
  year={2015},
  eprint={1502.05477},
  archivePrefix={arXiv}
}

@article{sutton1999options,
  title={Between MDPs and Semi-MDPs: A Framework for Temporal Abstraction in Reinforcement Learning},
  author={Sutton, Richard S. and Precup, Doina and Singh, Satinder},
  journal={Artificial Intelligence},
  volume={112},
  number={1--2},
  pages={181--211},
  year={1999},
  doi={10.1016/S0004-3702(99)00052-1}
}

@article{surana2026rollout,
  title   = {Generate, Filter, Control, Replay: A Comprehensive Survey of Rollout Strategies for {LLM} Reinforcement Learning},
  author  = {Surana, Rohan and Mundada, Gagan and Jiang, Xunyi and Wang, Chuhan and Tang, Zhenwei and Jiao, Difan and Huang, Zihan and Xiong, Yuxin and Wu, Junda and Yu, Sheldon and Li, Xintong and Jain, Raghav and Kuang, Nikki and Zhou, Sizhe and Jin, Bowen and Chu, Zhendong and Yu, Tong and Rossi, Ryan and Huang, Kuan-Hao and Shang, Jingbo and Han, Jiawei and McAuley, Julian},
  journal = {Transactions on Machine Learning Research (TMLR)},
  note    = {arXiv:2605.02913},
  year    = {2026}
}

@article{mundada2026wsgrpo,
  title   = {{WS-GRPO}: Weakly-Supervised Group-Relative Policy Optimization for Rollout-Efficient Reasoning},
  author  = {Mundada, Gagan and Huang, Zihan and Surana, Rohan and Yu, Sheldon and Zhang, Jennifer Yuntong and Li, Xintong and Yu, Tong and Yao, Lina and Shang, Jingbo and McAuley, Julian and Wu, Junda},
  journal = {arXiv preprint arXiv:2602.17025},
  year    = {2026}
}

@article{xiong2026rrpo,
  title   = {{RRPO}: Reference-Relative Policy Optimization with Stratified Conditional Rollouts},
  author  = {Xiong, Yuxin and Jiang, Xunyi and Surana, Rohan and Li, Xintong and Yu, Sheldon and Kuang, Nikki Lijing and Rossi, Ryan A. and Shang, Jingbo and Yu, Tong and McAuley, Julian and Wu, Junda},
  journal = {arXiv preprint arXiv:2607.18470},
  year    = {2026}
}

@article{surana2026fgrpo,
  title   = {{F-GRPO}: Factorized Group-Relative Policy Optimization for Unified Candidate Generation and Ranking},
  author  = {Surana, Rohan and Mundada, Gagan and Wu, Junda and Li, Xintong and Jiao, Yizhu and Jin, Bowen and Zhou, Sizhe and Yu, Tong and Sinha, Ritwik and Han, Jiawei and Shang, Jingbo and McAuley, Julian},
  journal = {arXiv preprint arXiv:2605.12995},
  year    = {2026}
}

@inproceedings{yu2025stateaware,
  title     = {Explainable Chain-of-Thought Reasoning: An Empirical Analysis on State-Aware Reasoning Dynamics},
  author    = {Yu, Sheldon and Xiong, Yuxin and Wu, Junda and Li, Xintong and Yu, Tong and Chen, Xiang and Sinha, Ritwik and Shang, Jingbo and McAuley, Julian},
  booktitle = {Findings of the Association for Computational Linguistics: EMNLP 2025},
  pages     = {16660--16667},
  year      = {2025}
}

@article{wei2026reasoningenergy,
  title   = {How Hard Does It Think? Analyzing Step-Aware Reasoning Energy in {LLM} Chain-of-Thought Trajectories},
  author  = {Wei, Hui and Wu, Junda and Yu, Sheldon and Zhou, Sizhe and Jiao, Yizhu and Zhong, Ming and Jin, Bowen and Yu, Tong and Pan, Shijia and Han, Jiawei and McAuley, Julian},
  journal = {arXiv preprint arXiv:2607.28674},
  year    = {2026}
}

@article{yu2026selfloops,
  title   = {Can We Break {LLMs} Out of Self-Loops? Fine-Grained Reasoning Control with Activation Steering},
  author  = {Yu, Sheldon and Yu, Tong and Jiang, Xunyi and Surana, Rohan and Mundada, Gagan and Kim, Sungchul and Yao, Lina and McAuley, Julian and Wu, Junda},
  journal = {arXiv preprint arXiv:2607.18100},
  year    = {2026}
}

@inproceedings{wu2025ctrls,
  title     = {{CTRLS}: Chain-of-Thought Reasoning via Latent State-Transition},
  author    = {Wu, Junda and Xiong, Yuxin and Li, Xintong and Yu, Sheldon and Hu, Zhengmian and Yu, Tong and Wang, Rui and Chen, Xiang and Shang, Jingbo and McAuley, Julian},
  booktitle = {Proceedings of the 29th International Conference on Artificial Intelligence and Statistics (AISTATS)},
  year      = {2026}
}

@inproceedings{wu2025ocean,
  title     = {{OCEAN}: Offline Chain-of-thought Evaluation and Alignment in Large Language Models},
  author    = {Wu, Junda and Li, Xintong and Wang, Ruoyu and Xia, Yu and Xiong, Yuxin and Wang, Jianing and Yu, Tong and Chen, Xiang and Kveton, Branislav and Yao, Lina and Shang, Jingbo and McAuley, Julian},
  booktitle = {International Conference on Learning Representations (ICLR)},
  year      = {2025}
}

@article{vishe2026skillr1,
  title   = {{Skill-R1}: Agent Skill Evolution via Reinforcement Learning},
  author  = {Vishe, Yash and Surana, Rohan and Jiang, Xunyi and Huang, Zihan and Li, Xintong and Kuang, Nikki Lijing and Yu, Tong and Rossi, Ryan A. and Shang, Jingbo and McAuley, Julian and Wu, Junda},
  journal = {arXiv preprint arXiv:2605.09359},
  year    = {2026}
}

@inproceedings{wang2025selfparam,
  title     = {Self-Updatable Large Language Models by Integrating Context into Model Parameters},
  author    = {Wang, Yu and Liu, Xinshuang and Chen, Xiusi and O'Brien, Sean and Wu, Junda and McAuley, Julian},
  booktitle = {International Conference on Learning Representations (ICLR)},
  year      = {2025}
}

@inproceedings{xia2025sand,
  title     = {{SAND}: Boosting {LLM} Agents with Self-Taught Action Deliberation},
  author    = {Xia, Yu and Shen, Yiran and Wu, Junda and Yu, Tong and Kim, Sungchul and Rossi, Ryan A. and Yao, Lina and McAuley, Julian},
  booktitle = {Proceedings of the 2025 Conference on Empirical Methods in Natural Language Processing (EMNLP)},
  year      = {2025}
}

@article{xie2025pluralistic,
  title   = {A Survey on Personalized and Pluralistic Preference Alignment in Large Language Models},
  author  = {Xie, Zhouhang and Wu, Junda and Shen, Yiran and Xia, Yu and Li, Xintong and Chang, Aaron and Rossi, Ryan and Kumar, Sachin and Majumder, Bodhisattwa Prasad and Shang, Jingbo and Ammanabrolu, Prithviraj and McAuley, Julian},
  journal = {arXiv preprint arXiv:2504.07070},
  year    = {2025}
}

@inproceedings{li2026importancedpo,
  title     = {Importance Sampling for Multi-Negative Multimodal Direct Preference Optimization},
  author    = {Li, Xintong and Wang, Chuhan and Wu, Junda and Surana, Rohan and Yu, Tong and McAuley, Julian and Shang, Jingbo},
  booktitle = {International Conference on Learning Representations (ICLR)},
  year      = {2026}
}

@article{surana2026massdpo,
  title   = {{MASS-DPO}: Multi-negative Active Sample Selection for Direct Policy Optimization},
  author  = {Surana, Rohan and Li, Xintong and Yu, Sheldon and Shen, Yiran Jenny and Wang, Chuhan and Yu, Tong and Ammanabrolu, Prithviraj and Shang, Jingbo and McAuley, Julian and Wu, Junda},
  journal = {arXiv preprint arXiv:2605.10784},
  year    = {2026}
}

@article{kveton2025activedpo,
  title   = {Active Learning for Direct Preference Optimization},
  author  = {Kveton, Branislav and Li, Xintong and McAuley, Julian and Rossi, Ryan and Shang, Jingbo and Wu, Junda and Yu, Tong},
  journal = {arXiv preprint arXiv:2503.01076},
  year    = {2025}
}

@article{wu2025irpo,
  title   = {In-context Ranking Preference Optimization},
  author  = {Wu, Junda and Surana, Rohan and Xie, Zhouhang and Shen, Yiran and Xia, Yu and Yu, Tong and Rossi, Ryan A. and Ammanabrolu, Prithviraj and McAuley, Julian},
  journal = {arXiv preprint arXiv:2504.15477},
  year    = {2025}
}

@article{zhou2026filesystem,
  title   = {Filesystem-Based Memory for {LLM} Agents: Organization, Evolution, and Sustainability},
  author  = {Zhou, Sizhe and Yu, Sheldon and Wei, Hui and Wu, Junda and Ouyang, Siru and Jiao, Yizhu and Pan, Shijia and McAuley, Julian and Zhang, Yu and Yu, Tong and Han, Jiawei},
  journal = {arXiv preprint arXiv:2607.26637},
  year    = {2026}
}

@article{hu2026skillbrew,
  title   = {{SkillBrew}: Multi-Objective Curation of Skill Banks for {LLM} Agents},
  author  = {Hu, Wentao and Chu, Zhendong and Zhang, Yiming and Wu, Junda and Jin, Ming and Zhao, Xiangyu and Shao, Yilei and Wang, Yanfeng and Wen, Qingsong},
  journal = {arXiv preprint arXiv:2605.29440},
  year    = {2026}
}

@inproceedings{yu2026olivia,
  title     = {{OLIVIA}: Online Learning via Inference-time Action Adaptation for Decision Making in {LLM} {ReAct} Agents},
  author    = {Yu, Sheldon and Wu, Junda and Li, Xintong and Kuang, Nikki Lijing and Zhou, Sizhe and Yu, Tong and Han, Jiawei and Shang, Jingbo and McAuley, Julian},
  booktitle = {Proceedings of the 2026 Conference on Empirical Methods in Natural Language Processing (EMNLP)},
  note      = {arXiv:2605.11169},
  year      = {2026}
}

\appendix

\section{Prompts}
\label{app:template}

Every prompt the method uses is listed here verbatim. All of them carry the same system prompt,
\texttt{You are an expert mathematician. Solve the problem carefully and precisely.} Curly braces
are the template's placeholders.

\paragraph{Solve prompt.} The prompt the policy acts on. The note block is present only in a
replay or in the teacher pass; the student and the deployed model never see it. On the science
sets one extra line is appended: \texttt{If the answer is a number, put only the number in the
box, without units.}
{\small\begin{verbatim}
Problem:
{question}

Hindsight note from an earlier attempt at this problem:
[{note}]
Use it if it helps.

Solve the problem. Think step by step, then give the final answer in \boxed{}.
\end{verbatim}}

\paragraph{Note prompt.} Written once per failed attempt. The attempt text is truncated to
6,000 characters (head and tail) when longer. The writer sees no reference answer or solution.
{\small\begin{verbatim}
You are reviewing one attempt at a math problem and writing a short note that
the same solver will read before its next attempt at this problem.

Problem:
{question}

Attempt:
{attempt}

Verdict: the attempt was incorrect.

Write one note that names the core mistake and what to do instead, so the
solver avoids it.

Constraints:
- The note is advice to the solver, written as instructions, not a description
  of what happened.
- At most 80 words.
- Return ONLY valid JSON with exactly these fields:
{"summary": "one sentence on what the attempt did", "note": "the advice"}
\end{verbatim}}

\paragraph{Revision prompt.} Used once, for a note whose replays did not beat the group's
no-note rate. Up to three failed replays are shown, each truncated to 1,200 characters.
{\small\begin{verbatim}
You previously wrote a note for the problem below. When the solver used the
note, it was correct on {successes} of {total} new attempts.

Problem:
{question}

Previous note:
[{note}]

Excerpts from attempts that still failed with the note:
{failures}

Revise the note so that it prevents these failures. Keep what worked.

Constraints:
- The note is advice to the solver, written as instructions, not a description
  of what happened.
- At most 80 words.
- Return ONLY valid JSON with exactly these fields:
{"summary": "one sentence on why the previous note fell short",
 "note": "the revised advice"}
\end{verbatim}}

\paragraph{Answer-matching prompt (evaluation only).} On the two free-form test sets, an answer
the exact verifier rejects is shown to the evaluated model itself, decoded greedily with Qwen3's
\texttt{/no\_think} switch, and counted as correct if the reply is \texttt{Yes}. This prompt is
never used in training.
{\small\begin{verbatim}
You are grading a free-form answer to a question. Decide whether the candidate
answer means the same thing as the reference answer. Differences in wording,
units, formatting, rounding within about 1%, or extra explanation do not
matter; a different value, entity or claim does.

Question:
{question}

Reference answer:
{reference}

Candidate answer:
{candidate}

Does the candidate answer match the reference answer? Reply with exactly one
word: Yes or No. /no_think
\end{verbatim}}
\section{Implementation Details}
\label{app:impl}

\paragraph{Trainer.}
\method{} is a subclass of TRL's \texttt{GRPOTrainer}. TRL owns rollout generation, weight
synchronization to vLLM, distributed training, checkpoints, and the GRPO loss; the subclass adds
two hooks. After TRL has sampled and scored a batch, the rollouts of all training processes are
gathered on the main process, which runs the note search (write, replay, judge, revise) against
the shared vLLM server. Because TRL syncs the weights to that server every step, notes and replays
are always produced by the policy that produced the rollouts. The kept note and one replay per
failed rollout are then broadcast back to the process that owns the rollout. The reward is a single
function, one if the verifier accepts the completion's final answer and zero otherwise.

\paragraph{The distillation term.}
For every rollout with a kept note, the chosen replay is scored twice with the current weights:
once under the note prompt without gradient (the teacher) and once under the plain prompt with
gradient (the student). The student pass rides along in TRL's own loss forward as extra batch
rows, left-padded to a common width, rather than in a second forward pass, because ZeRO-2 reduces
each parameter's gradient once per backward and rejects a parameter used by two graphs. The
per-token weight is the clipped positive part of the teacher--student log-probability gap; the
term is the weighted negative log-likelihood averaged over the active replay tokens of the
micro-batch, scaled by the inverse of the gradient-accumulation count exactly as TRL scales its
own loss, and added with coefficient $\lambda$. When no rollout in the batch has a kept note on any
process, no rows are appended and the step is plain GRPO to the byte.

\paragraph{The committed carrier.}
The codebase also implements the alternative the paper argues against, and it produced the
middle curve of Figure~\ref{fig:curves}: the failed rollout's own tokens are re-scored under the
note, and the student is pushed up on them with weight $\sigma(5\,(\text{teacher}-\text{student}))$
per token. That run used the earlier configuration of the pipeline (coefficient $0.01$, KL
$0.01$, notes for successful rollouts too, no leak filter) and its final model scores $78.6$ on
MATH-500 under the protocol of Section~\ref{sec:experiments}, against $76.9$ for GRPO and $84.5$
for \method{}. A configuration-matched comparison is not yet run.

\begin{table}[ht]
\centering
\caption{Registered training settings, shared by every arm and backbone. All runs are full-parameter
(bf16, DeepSpeed ZeRO-2, gradient checkpointing) on 7 training GPUs with a vLLM server on the eighth.
GRPO-only baselines differ from \method{} only in the rows marked
$^\ast$. Nothing is fit besides the policy.}
\label{tab:hyper}
\small
\begin{tabular}{lll}
\toprule
Symbol/config & Meaning & Value \\
\midrule
\multicolumn{3}{l}{\emph{GRPO backbone}} \\
$n$ & rollouts per prompt & $8$ \\
 & problems per optimizer step & $14$ ($4$ rows/GPU $\times$ $4$ micro-steps $\times$ $7$ GPUs) \\
 & optimizer steps & $150$ (about one pass over 2,100 problems) \\
$\epsilon_{\mathrm{clip}}$ & ratio clip & $0.2$ \\
 & loss aggregation & token-level (DAPO) \\
$\beta_{\mathrm{KL}}$ & reference-model KL coefficient & $0$ (\method{}); $0.01$ (GRPO baseline)$^\ast$ \\
 & learning rate, warm-up & $10^{-6}$, none \\
 & sampling & temperature $1.0$, top-$p$ $1.0$ \\
 & prompt / completion budget & $1{,}024$ / $3{,}072$ tokens \\
\midrule
\multicolumn{3}{l}{\emph{Notes and replays}} \\
 & which rollouts get a note & failed rollouts of groups not solved $8/8$ \\
 & note writer & the policy itself, greedy, $2{,}048$-token budget, $\le 80$ words \\
$k$ & replays per note & $2$ (temperature $1.0$) \\
 & keep rule & replay success rate $>$ the group's no-note rate \\
 & revision rounds & $1$ (a rejected note is rewritten once and re-tested) \\
 & answer-leak filter & notes that state the reference answer are rejected \\
\midrule
\multicolumn{3}{l}{\emph{Distillation term}} \\
$\lambda_{\mathrm{OPD}}$ & auxiliary coefficient & $0.2$ \\
 & target & one replay of the kept note, chosen uniformly, imitated from the plain prompt \\
$w_t$ & per-token weight & $\mathrm{clip}\big((\log\pi(\tilde y_t\mid x\oplus s)-\log\pi(\tilde y_t\mid x))_+,\,0,\,2\big)$ \\
 & rows & failed rollouts only; empty when no note is kept (plain GRPO) \\
\bottomrule
\end{tabular}
\end{table}

\begin{table}[t]
\centering
\caption{\textbf{The note search in numbers} (Qwen3-4B on MATH, all 150 steps). Three notes in
four are discarded, and the two screens cost no replays at all. A step holds 14 problems and 112
rollouts.}
\label{tab:notes}
\small
\begin{tabular}{lr}
\toprule
\multicolumn{2}{l}{\emph{The filter} --- one note per failed rollout} \\
\midrule
notes written                          & 4,680 \\
\quad returned no usable note          & 310 \quad (6.6\%) \\
\quad stated the reference answer      & 517 \quad (11.0\%) \\
\quad replays did not beat the group   & 2,705 \quad (57.8\%) \\
\quad \textbf{kept}                    & \textbf{1,148} \quad (\textbf{24.5\%}) \\
\midrule
\multicolumn{2}{l}{\emph{What survives}} \\
\midrule
replay success rate, all judged notes  & 0.26 \quad (no note: 0.17) \\
kept notes that are revisions          & 33\% \\
kept notes from all-fail groups        & 17\% \\
distilled replays verified correct     & 79\% \\
replay tokens the note did not raise   & 56\% \\
\bottomrule
\end{tabular}
\end{table}

\paragraph{Note search details.}
Notes are written for failed rollouts of groups not solved by all $n$ rollouts. The writer is the
policy itself, decoded greedily with a 2,048-token budget, of which most is the model's own
thinking; a note whose completion hits the budget before producing a JSON object, or whose output
contains no note, is discarded without replay. The leak check applies the verifier's answer-match
routine to the note text (the normalized reference inside a boxed expression, or as a
whole-number match in the normalized text). Replays use the rollout sampling settings
(temperature $1.0$, 3,072 tokens). The keep rule compares the replay success rate with the
problem's no-note rate computed over all $n$ rollouts across processes. Notes that fail the rule
are revised once with the revision prompt of Appendix~\ref{app:template}; the revised note's own
replays are the ones distilled if it is kept. Per step the trainer logs every note with its verdict
and tally, and every tenth step it dumps full rollouts, notes, and replays for a few problems.

\paragraph{Evaluation.}
Trained checkpoints are evaluated with vLLM (tensor parallel 4, bf16), 8 samples per problem at
temperature $0.6$, top-$p$ $0.95$, and 4,096 tokens, and graded by the exact verifier of the set:
the normalized last boxed expression for MATH-500, and a number within 1\%, a list, or a
short-string match for the science sets. On the two free-form sets, every answer the exact
verifier rejects is additionally judged by the evaluated model itself with the answer-matching
prompt of Appendix~\ref{app:template}, and the tables report both numbers as exact\,/\,matched.
\section{Notation Summary}
\label{app:notation}

\begin{table}[ht]
\centering
\small
\begin{tabular}{ll}
\toprule
Symbol & Meaning \\
\midrule
\multicolumn{2}{l}{\emph{Rollouts and rewards}}\\
$x,\ g$ & prompt and its reference answer \\
$y=(y_1,\ldots,y_T)$, $y_{<t}$ & a completion and its prefix before token $t$ \\
$\{y_i\}_{i=1}^n$ & the GRPO group of $n$ rollouts for a prompt \\
$R(x,y)\in\{0,1\}$, $R_i$ & verifier reward (final answer matches $g$); reward of rollout $i$ \\
$F=\{i:R_i{=}0\}$, $\hat p$ & the failed set; the group's no-note success rate $\tfrac1n\sum_iR_i$ \\
$\pi_\theta,\ \pi_{\theta_{\old}}$ & the policy being trained; the frozen snapshot that sampled the group \\
\midrule
\multicolumn{2}{l}{{\emph{Finite-horizon verifier MDP}}}\\
{$h_t,\ a_t,\ H$} & {reasoning state, next action, completion budget} \\
{$P_\pi^x,\ G_x,\ J_x(P)$} & {trajectory law, success set, probability $P(G_x)$} \\
{$d_t^\pi$} & {distribution of states reached at time $t$} \\
{$V_t^\pi,\ Q_t^\pi$} & {expected terminal reward from a state; after an action} \\
{$A_t^\pi,\ V_t^\star$} & {state-action advantage; best feasible continuation value} \\
\midrule
\multicolumn{2}{l}{\emph{Notes and replays}}\\
$s,\ s_i$ & a hindsight note; the note written about failed rollout $i$ ($\varnothing$ = none kept) \\
$x\oplus s$ & the solve prompt with the note block inserted \\
$k$, $\tilde y^{(j)}$ & replays per note; the $j$-th continuation sampled under the note \\
$\hat r(s)$ & replay success rate $\tfrac1k\sum_jR(x,\tilde y^{(j)})$; kept iff $\hat r(s)>\hat p$ \\
$\tilde y_i$ & the replay chosen (uniformly) for distillation \\
\midrule
\multicolumn{2}{l}{\emph{Distillation term}}\\
{$\teacher=\sg[\pi_{\theta_{\old}}](\cdot\mid x\oplus s)$} & {stop-gradient snapshot teacher reading the note} \\
$w_{i,t}$, $c$ & per-token lift weight $\clip(\log\teacher(\tilde y_{i,t})-\log\pi_\theta(\tilde y_{i,t}\mid x,\cdot),0,c)$; its cap \\
$\mathcal A$ & the set of credited replay tokens in a batch \\
$\loss_{\mathrm{GRPO}},\ \loss_{\method},\ \lambda$ & GRPO loss; the distillation term; its coefficient \\
$\sg[\cdot]$ & stop-gradient \\
\bottomrule
\end{tabular}
\end{table}

\section{Formal Setup, Proofs, and Boundary Cases}
\label{app:exploration-theory}

All trajectory laws, states, rewards, and values below use the verifier MDP of
Definition~\ref{def:reasoning-mdp}. We condition on a fixed problem $x$ throughout.
This appendix states the no-reward-contrast failure of Section~\ref{sec:exploration-theory}
formally, proves the results stated there, and records their boundary cases.

\begin{proof}[Proof of Lemma~\ref{lem:allfail-signal}]
Independence gives the probability. On $\mathcal E_{\mathrm{fail}}$,
both the sample reward mean and its
standard deviation are zero, so each advantage
$(R_i-\bar R)/(\sigma_R+\varepsilon)$ vanishes. Every clipped
task-reward-surrogate term is therefore identically zero in the new policy parameters.
KL regularization and gradients from other prompts are not included in this claim.
\end{proof}

\begin{proof}[Proof of Proposition~\ref{prop:repair-ceiling}]
A successful trajectory must be outside $\mathcal B(y^-)$. Hence the chance that retry $j$
succeeds, conditional on all previous retries failing, is at most $\epsilon$. Repeated
conditioning yields $\Pr(\text{no success in }m\text{ retries})\ge(1-\epsilon)^m$;
Bernoulli's inequality gives $1-(1-\epsilon)^m\le m\epsilon$.
Independence between retries is unnecessary.
\end{proof}

For a concrete witness, take a horizon-$3$ process whose first decision selects strategy
$A$ or $B$. After $A$, three further individually valid transitions are required for
success; after $B$, two suffice. The prefix committing to $A$ has $V_1^\star(h_1)=0$ with
the remaining budget, although none of its transitions is locally invalid. Any repair
kernel preserving that decision has $\epsilon=0$. An unconstrained restart can choose
$B$ and is therefore outside the hard-ceiling hypothesis. More generally, if the
conditional chance of leaving the basin is not uniformly small, the soft bound says
little about a full-restart reflection method.

\begin{proof}[Proof of Proposition~\ref{prop:replay-selection}]
Conditional on the original history and fixed $s$, the $k$ replay outcomes are
independent Bernoulli variables with parameter $p_s$. They all fail with probability
$(1-p_s)^k$. Choose an index $U$ uniformly, independently of those outcomes.
The event $\{R(x,Z_U)=1\}$ implies admission and has unconditional probability $p_s$,
so conditioning on admission gives
$J_x(Q_s)=p_s/[1-(1-p_s)^k]$. This is at least $p_s$ since its
denominator is at most one. The function $1-(1-u)^k$ increases strictly with
$u\in[0,1]$. Compare $u=p_s$ to $p_0$ for plain retries, or to the bound
$u=\epsilon$ in Proposition~\ref{prop:repair-ceiling}.
\end{proof}

The acceptance event selects a batch enriched in successful outcomes; it does not
certify the content of its note. In particular, $p_s=p_0>0$ can yield admission
after the original group happens to fail $n$ times. A uniformly chosen accepted
replay is not guaranteed correct when $k>1$. Adaptive proposals and revisions
require conditioning on the history \emph{before} each new replay batch. The
same-attempt comparison also excludes note-writing tokens from its cost.

\begin{lemma}[Finite-horizon performance difference]
\label{lem:performance-difference}
For two plain-prompt policies $\pi,\pi'$ on the verifier MDP, let
$A_t^\pi(h,a)=Q_t^\pi(h,a)-V_t^\pi(h)$ be the terminal-reward advantage
and $d_t^{\pi'}$ the state law under $\pi'$ at time $t$. Then
\begin{equation*}
J_x(P_{\pi'}^x)-J_x(P_\pi^x)
=\sum_{t=0}^{H-1}
  \E_{h_t\sim d_t^{\pi'},\,a_t\sim\pi'(\cdot\mid h_t)}
  [A_t^\pi(h_t,a_t)].
\end{equation*}
Thus a nonnegative expected $\pi$-advantage at every state visited by $\pi'$
is sufficient for non-degradation.
\end{lemma}
\begin{proof}
Set $V_H^\pi(h_H)=R(x,y)$ for padded trajectories. Conditional on
$(h_t,a_t)$ under $\pi'$, the expected difference
$V_{t+1}^\pi(h_{t+1})-V_t^\pi(h_t)$ is $A_t^\pi(h_t,a_t)$.
Summing over time telescopes to
$\E_{\pi'}R-V_0^\pi(h_0)=J_x(P_{\pi'}^x)-J_x(P_\pi^x)$.
\end{proof}

This is the finite-horizon performance-difference identity underlying conservative
policy iteration. A better hinted policy does not establish nonnegative advantages
for a plain-prompt student.

\begin{proof}[Proof of Theorem~\ref{thm:local-transfer}]
For the mixture of complete-trajectory laws,
\begin{equation*}
J_x(M_\alpha)
=(1-\alpha)J_x(P_{\mathrm G})+\alpha J_x(Q)
\ge J_x(P_{\mathrm G})+\alpha\delta.
\end{equation*}
Because $G_x$ is an event,
$J_x(P_{\mathrm A})\ge J_x(M_\alpha)
-\operatorname{TV}(P_{\mathrm A},M_\alpha)$.
The asserted inequality and non-degradation condition follow.
\end{proof}

Pinsker's inequality supplies a stronger, distributional sufficient condition:
\begin{equation*}
\KL(M_\alpha\Vert P_{\mathrm A})\le 2(\alpha\delta)^2
\quad\Longrightarrow\quad
\operatorname{TV}(P_{\mathrm A},M_\alpha)\le\alpha\delta.
\end{equation*}
For padded autoregressive trajectories of common horizon, the KL chain rule is
\begin{equation*}
\KL(M_\alpha\Vert P_{\mathrm A})
=\sum_{t=0}^{H-1}\E_{h_t\sim M_\alpha}
  \KL\!\left(M_\alpha(\cdot\mid h_t)
           \,\Vert\,\pi_{\mathrm A}(\cdot\mid h_t)\right).
\end{equation*}
Thus per-prefix distributional distillation could establish the transfer premise
if it matches the \emph{mixture's} conditionals on the \emph{mixture's} prefixes.
Neither clipping a GRPO ratio nor optimizing the selected-token positive-lift
loss of Section~\ref{sec:method-objective} alone certifies this KL bound.
The mixture parameter $\alpha$ is not automatically the auxiliary loss coefficient;
the theorem is for one problem and one local update, not an entire shared-parameter run.

\begin{proof}[Proof of Proposition~\ref{prop:strategy-lift}]
For a selected $z=0$, $w_0=0$ and $\phi$ is unchanged. For $z=1$,
$\partial_\phi\log\pi_\phi(1)=1-p$, so gradient descent raises the logit
to $\phi+\eta\lambda w_1(1-p)$. The sigmoid is strictly increasing,
and Proposition~\ref{prop:replay-selection} yields
$\Pr(z=1\mid E_s)=\mu>0$. Lemma~\ref{lem:allfail-signal}
gives the GRPO assertion.
\end{proof}

\end{document}